\def\year{2022}\relax
\documentclass[letterpaper]{article} 
\usepackage{aaai22}  
\usepackage{times}  
\usepackage{helvet}  
\usepackage{courier}  
\usepackage[hyphens]{url}  
\usepackage{graphicx} 
\usepackage{natbib}  
\usepackage{caption} 
\DeclareCaptionStyle{ruled}{labelfont=normalfont,labelsep=colon,strut=off} 
\usepackage{soul}
\usepackage{url}
\usepackage[hidelinks]{hyperref}
\usepackage[utf8]{inputenc}
\usepackage{amsmath}
\usepackage{amsthm}
\usepackage{booktabs}
\usepackage{algorithm}
\usepackage{cite}
\usepackage[caption=false]{subfig}
\usepackage{amsmath}
\usepackage{multirow}
\usepackage{amssymb}
\usepackage{amsfonts}
\usepackage{algpseudocode}
\usepackage{rotating}
\usepackage{tikz}
\usetikzlibrary{shapes,arrows,arrows.meta}
\usepackage{easyReview}
\usepackage{comment}
\usepackage{multirow}
\usepackage{multicol}
\usepackage{booktabs}
\usepackage{xspace}

\usepackage{newfloat}
\usepackage{listings}
\floatstyle{ruled}
\newfloat{listing}{tb}{lst}{}
\floatname{listing}{Listing}

\newcommand{\red}[1]{\textcolor{red}{#1}}

\newcommand{\new}[1]{\textcolor{black}{#1}}

\title{CrossWalk: Fairness-enhanced Node Representation Learning}
\author {
    Ahmad Khajehnejad,\textsuperscript{\rm 1}
    Moein Khajehnejad,\textsuperscript{\rm 2}
    Mahmoudreza Babaei,\textsuperscript{\rm 3}
    Krishna P. Gummadi,\textsuperscript{\rm 4}
    Adrian Weller,\textsuperscript{\rm 5,6}
    Baharan Mirzasoleiman \textsuperscript{\rm 7}
}
\affiliations {
    \textsuperscript{\rm 1} Dept of Computer Engineering, Sharif University of Technology\\
    \textsuperscript{\rm 2} Dept of Data Science and AI, Monash University\\
    \textsuperscript{\rm 3} Max Planck Institute for Human Development\\
    \textsuperscript{\rm 4} Max Planck Institute for Software Systems\\
    \textsuperscript{\rm 5} University of Cambridge\\
    \textsuperscript{\rm 6} The Alan Turing Institute\\
    \textsuperscript{\rm 7} Dept of Computer Science, University of California Los Angeles\\
    khajenezhad@ce.sharif.edu,
    moein.khajehnejad@monash.edu,  
    babaei@mpib-berlin.mpg.de
}
\usepackage{bibentry}

\begin{document}

\maketitle

\begin{abstract}
The potential for machine learning systems to amplify social inequities and unfairness is receiving increasing popular and academic attention. Much recent work has focused on developing algorithmic tools to assess and mitigate such unfairness. However, there is little work on enhancing fairness in graph algorithms. Here, we develop a simple, effective and general method, CrossWalk, that enhances fairness of various graph algorithms, including influence maximization, link prediction and node classification, applied to node embeddings. CrossWalk is applicable to any random walk based node representation learning algorithm, such as DeepWalk and Node2Vec. The key idea is to bias random walks to cross group boundaries, by upweighting edges which (1) are closer to the groups' peripheries or (2) connect different groups in the network. CrossWalk pulls nodes that are near groups' peripheries towards their neighbors from other groups in the embedding space, while preserving the necessary structural properties of the graph. Extensive experiments show the effectiveness of our algorithm to enhance fairness in various graph algorithms, including influence maximization, link prediction and node classification in synthetic and real networks, with only a very small decrease in performance. 
\end{abstract}

\section{Introduction}

\label{sec:introduction}

Fairness in machine learning is receiving growing attention as algorithmic systems are increasingly deployed across society in ways that have significant impact on people's lives. 
Decisions made by such systems often affect different population subgroups disproportionately \cite{kirchner2016machine, osoba2017intelligence}. As a result, 
fairness --- the absence
of 
prejudice or favoritism toward an individual or a group based on their inherent or acquired
characteristics --- has received much recent interest. 
See \cite{mehrabi2019survey} for a recent survey.

Despite the emerging body of efforts to enhance algorithmic fairness in machine learning algorithms, 
there has been little work on enhancing fairness in graph algorithms. 
Considering the far-reaching application of graphs to many important problems in sociology, finance, computer science, and operations research \cite{easley2010networks}, it becomes crucial to develop methods to enhance fairness in graph algorithms, to prevent outcomes skewed for or against a particular group of people.
A concrete example is promoting fairness of influence maximization 
in social networks, which aims ultimately to reach different population groups roughly equally. This becomes critical for receiving important information, such as job opportunities 
or loan advertisements, by various communities. 
Another example is enhancing fairness in node classification in social networks, which aims to limit disparity in the prediction accuracy across different groups or social communities. 

Here, we develop a simple, intuitive and effective approach, CrossWalk, to promote fairness in the results of graph algorithms applied to node embeddings produced by random walk-based node representation learning algorithms.
%
The key idea of our method is to 
bias random walks to cross group boundaries,
by upweighting edges which are (1) closer to the groups' peripheries
or (2) connecting different groups in the network. 
Several methods for  representation learning on graphs, including DeepWalk \cite{deepwalk} and Node2Vec \cite{grover2016node2vec}, leverage random walks to preserve higher-order proximity between nodes and learn the corresponding representations.
When transition probabilities in random walks are chosen by CrossWalk, 
random walks initiated from a particular group will be pulled toward the group boundaries and have a higher probability of crossing groups' peripheries and visiting nodes from other groups in the network. In doing so, 
CrossWalk pulls closer nodes that are near groups’ peripheries towards their neighbors from other groups in the embedding space, while preserving the necessary structural information from the graph.
The resulting representation 
promotes fairness for the original graph in the result of various graph algorithms,  
including influence maximization, node classification, and link prediction.
%
Beyond representation learning, our method is applicable to any graph algorithm which works by stochastic traversal of network edges, such as the classical influence maximization based on the Independent Cascade (IC) model, which we discuss in Section \ref{sec:appendix-classical-IM} of Supplementary Material.

We conduct experiments on synthetic and real-work networks to evaluate the effectiveness of our proposed method. We first apply our method to learn node representations, using DeepWalk and Node2Vec 
in a number of synthetic networks, as well as two real-world networks, namely, Rice-Facebook \cite{mislove2010you} and a subset of Twitter \cite{babaei2016efficiency,cha_icwsm10}. We then apply various graph algorithms, including influence maximization, node classification, and link prediction to the obtained node representations.  
We show that CrossWalk is very effective in enhancing fairness of the aforementioned graph algorithms in synthetic and real networks, without a significant compromise in the total performance of the algorithms.

\section{Related Work}
There has been much effort recently to enhance fairness in machine learning algorithms.
However, few works have focused on detecting and mitigating unfairness in graph algorithms such as influence maximization, node classification, and link prediction. 
In this section, we review recent
works on network embedding, and fairness in influence maximization, node classification, and link prediction. 

\vspace{-1mm}
\subsection{Representation Learning on Graphs}

Node representation learning algorithms attempt to map the nodes in a graph to a lower dimensional space, such that the network structure is preserved. 
Among the most well-known approaches are embedding methods based on random-walks \cite{khajehnejad2019simnet,grover2016node2vec}, deep learning architectures \cite{wang2016structural}, and graph neural networks \cite{hamilton2017inductive}.  
%
Most related methods to our work are DeepWalk and Node2Vec, that are two widely used random walk based methods for deriving node representations. 

\vspace{1mm}
\noindent\textbf{DeepWalk}
DeepWalk \cite{deepwalk} takes a graph $G$ and iteratively (1) initiates a random walk from a randomly sampled vertex, and (2) updates the node representations, by optimizing the Skip-gram likelihood objective \cite{mikolov2013efficient}, using a hierarchical soft-max. 
DeepWalk preserves higher-order proximity between nodes by maximizing the probability of observing the last $d$ nodes and the next $d$ nodes in the random walk centered at $v_i$, i.e. maximizing $\log P(v_{i-d}, \cdots, v_{i-1},\allowbreak v_{i+1}, \cdots, v_{i+d}|\Phi_i)$, where $2d+1$ is the length of the random walk and $\Phi_i$ is the representation of node $v_i$.
The model 
performs the optimization over sum of log-likelihoods for each random walk.

\vspace{1mm}\noindent\textbf{Node2Vec}
Similar to DeepWalk, Node2Vec \cite{grover2016node2vec} preserves higher-order proximity between nodes by maximizing the probability of occurrence of subsequent nodes in fixed length random walks. The crucial difference from DeepWalk is that Node2Vec employs biased-random walks that provide a trade-off between breadth-first (BFS) and depth-first (DFS) graph searches, and hence produces higher-quality and more informative embeddings than DeepWalk. Choosing the right balance enables Node2Vec to preserve community structure and structural equivalence between nodes.

\vspace{-1mm}
\subsection{Fair Influence Maximization}
Influence maximization aims at
finding an initial set of nodes to maximize the number of further adopters \cite{richardson2002mining}. 
%
While finding the optimal solution for influence maximization is NP-hard \cite{kempe2003maximizing}, a simple greedy algorithm provides a $1-1/e$-approximation guarantee
under different 
cascade models such as Linear Threshold (LT) and Independent Cascade (IC). 
Since then, extensive research has focused on studying different variations \cite{goyal2013minimizing,carnes2007maximizing} among which \citet{keikha2020influence} take advantage of a network embedding approach by applying a $k$-means method on the embedding space to select the resulting $k$ cluster centroids as initial seeds. 

Among the wide range of recent works on influence maximization, only a few has considered the fairness. 
After 
\citet{babaei2016efficiency} showed that users in social media select their sources sub-optimally in the sense of receiving 
diverse information, a group of studies have worked on diversifying the initial seeds 
\cite{benabbou2018diversity,aghaei2019learning}. However, these works still fail to take into account the fairness criterion. 
Network embedding approaches for fair influence maximization which have been introduced recently, 
\citet{khajehnejad2020adversarial}
proposed adversarial network representation learning for enhancing fairness of influence maximization. We use this method as a baseline in our work.

\subsection{Fair Node Classification}
Node classification determines the labelling of nodes by looking at the neighbours' labels.
Recent node classification methods work by classifying the learned nodes representations. 
A few studies considered finding fair representations for classification 
\cite{zemel2013learning,lahoti2019operationalizing}. 
The key idea of the above work is that individuals who are deemed similar according to a task-specific similarity metric should receive similar outcomes.
However, we are not aware of any fariness-enhanced node representation learning algorithm for classification in graphs. 

\subsection{Fair Link Prediction}
Link prediction infers new or previously unknown relationships of a network. However, existing algorithms are susceptible to promoting links
that may lead to increased segregation. 
Among the few existing methods to enhance fairness of link prediction, FairWalk \cite{rahman2019fairwalk} is a modified random walk, which results in a more diverse network neighborhood representation thereby producing less biased graph embedding. FairWalk, however, fails to enhance fairness in graphs where the majority of nodes are more than one hop away from group peripheries.
In addition, while FairWalk was only applied to friendship recommendation in \cite{rahman2019fairwalk}, in this paper we show the applicability of CrossWalk to a wider class of graph algorithms: influence maximization, link prediction, and node classification.
%
More recently, methods that combine adversarial network representation learning with supervised link prediction to enhance fairness of link prediction were proposed in \cite{masrour2020bursting, bose2019compositional},
and a Bayesian method which utilizes a biased prior in the embedding phase to generate fair node representations was proposed in \cite{buyl2020debayes}. 

Existing methods to enhance fairness of node representation learning are specific to a specific graph algorithm. On the other hand, we propose a general and effective method to enhance fairness of node representation learning that is applicable to various graph algorithms.
While FairWalk \cite{rahman2019fairwalk} was originally proposed to enhance fairness of link prediction, we use it as a baseline for various graph algorithms in our work.




\section{Problem Formulation}

Consider a directed network $G=(V,E)$ with a set 
$V$ of nodes and a set $E$ of edges. We denote by $w_{uv}\in\mathbb{R}$ the weight of an edge $(u,v)\in E$.  Assume the nodes are partitioned into $C$ groups $\{V_1, \cdots, V_C\}$. Furthermore, let $l_v$ indicate the group that node $v$ belongs to, and $\mathcal{N}(v)=\{u|(v,u)\in E\}$ be the set of nodes in $v$'s immediate neighborhood.
%
Many algorithms for representation learning on graphs, including DeepWalk and Node2Vec, leverage random walks to preserve higher-order proximity between nodes and learn their 
representations. 

\vspace{2mm}\noindent\textbf{Random Walk:}
Given a source node $u_i$, a random walk $\mathcal{W}_{u_i}$ of length $d$ rooted at $u_i$ is a sequence of vertices $u_i, u_{i+1}, \cdots , u_{i+(d-1)}$, not necessarily distinct, such that $(u_i, u_{i+1})$ is an edge in the graph. 
Formally, nodes $u_i$ are generated according to the following distribution:
\begin{equation}
    P(u_i=u|u_{i-1}=v)= 
\begin{cases}
    \pi_{vu}
    & if (i,j)\in E\\
    0,              & \text{otherwise}
\end{cases}
\vspace{-.5mm}\end{equation}
where $\pi_{ij}$ is the normalized transition probability between nodes $i$ and $j$.
In a weighted network, normalized edge weights $w_{uv}$ can be used as transition probabilities $\pi_{uv}$ of the random walk.

\begin{table}[t]
\centering
\caption{Notations used in this paper}

\label{tab:notations}
\scalebox{0.65}{
\begin{tabular}{c  p{10cm} l } \toprule
\textbf{Notation} & \textbf{Definition} &  \\ \midrule
$G$ & Original Graph \\ \midrule
$V$, $E$ & Nodes and edges of $G$ \\ \midrule
$V_i$ & The $i^{th}$ group of $G$ \\ \midrule
$S$ & Initial seed set for IC model \\ \midrule
$Q$ & Performance of a graph algorithm on the entire graph
\\ \midrule
$Q_i$ & 
Performance of a graph algorithm on group $i\in [C]$
\\ \midrule
$\pi_{uv}$& Normalized transition probability between nodes $u$ and $v$ \\ \midrule
$\mathcal{W}_{u}$ & A random walk with fixed length $d$ rooted at node $u$\\ \midrule
$m(u)$& Closeness of node $u$ to group boundaries\\ \midrule
$\mathcal{N}(u)$& The set of neighbors of node $u$\\ \midrule
$l_u$& The group to which node $u$ belongs\\ \midrule
$N_u$& The set of $u$'s neighbors within $l_u$ \\ \midrule
$R_u$& The set of $u$'s neighbors within groups other than $l_u$ \\ \midrule
$N^c_v$& The set of $u$’s neighbors that belong to another group $c=l_v \neq l_u$ \\ \midrule
 $p$ & Tuning parameter controlling how much information propagates from one group to the others. \\ \midrule
 $\alpha$ & Multiplication factor in the reweighting process controlling the probability of a boundary node being connected to other groups.   \\ \midrule
 $w_{uv}$& Edge weight between nodes $u$ and $v$ in the original graph \\ \midrule
 $w'_{uv}$&  New edge weight between nodes $u$ and $v$ after reweighting method \\ \bottomrule
\end{tabular}
}
\end{table}

\vspace{2mm}\noindent\textbf{Fairness Metric (Disparity):}
Let $Q\in\mathbb{R}$ be the performance of algorithm $A$ on the entire graph, 
and $Q_i\in\mathbb{R}$
be the performance of $A$
on group $i\in[C]$ in the graph.
Our goal is to modify the weight $w_{uv}$ of every edge $(u,v)\in E$ to $w'_{uv}$ so that when used by random walks to produce node embedding, the performance of $A$ applied to the produced representations has a higher fairness and smaller discrepancy on different groups of the 
underlying graph. More formally, we will propose a mapping from the original edge weights $\{w_{uv}\}$ to modified weights $\{w'_{uv}\}$ which will be used by random walks to produce node representations. The aim is that $A$ applied to node embeddings
achieves a low value of 
\begin{equation} \label{eq:fairness-def}
    \text{disparity}(A) = Var(\{Q_i\}:i\in [C])
\end{equation}
\new{In most learning applications, the output of a model can be represented as a set of decisions on some items each of which belongs to a group, and $Q$ is the fraction of positive decisions. For example, in the problem of influence maximization, $Q$ is the fraction of nodes that are infected at the end of a network diffusion process, and $Q_i$ is the fraction of infected nodes in group $i$. In node classification/link prediction, $Q$ is the fraction of correctly labeled/detected nodes/links, and $Q_i$ is the fraction of correctly labeled/detected nodes within group $i$ (for link prediction we can assume cross group edges as some extra groups as well). In all of these cases, minimising the disparity measure defined in equation \eqref{eq:fairness-def} is equivalent to reducing the dependency between the model's decision and the sensitive attribute, and can be considered as a from of demographic parity \cite{hardt2016equality}.}

\section{Our Method: CrossWalk}
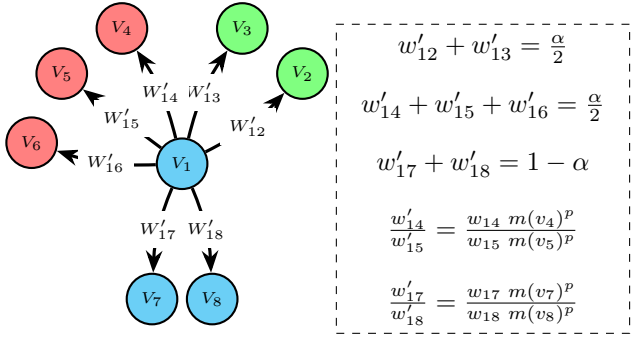
\begin{figure}[!htb]
\vspace{-2mm}
    \begin{center}
        \def \sc {0.2}
\begin{tikzpicture}[scale=\sc]

\def \Ox {0}
\def \Oy {0}
\def \A {A}
\def \fs {\tiny}

\node (rect2) at (20+\Ox,-1+\Oy) [draw,dashed,minimum width=\sc*8cm,minimum height=\sc*5cm] {\begin{tabular}{c} \( w'_{12}+w'_{13} = \frac{\alpha}{2} \) \\ \\
\( w'_{14}+w'_{15}+w'_{16} = \frac{\alpha}{2} \) \\ \\
\( w'_{17}+w'_{18} = 1-\alpha \) \\ \\
\( \frac{w'_{14}}{w'_{15}} = \frac{w_{14}~m(v_4)^p}{w_{15}~m(v_5)^p} \) \\ \\
\( \frac{w'_{17}}{w'_{18}} = \frac{w_{17}~m(v_7)^p}{w_{18}~m(v_8)^p} \)
\end{tabular} };

\begin{scope}[every node/.style={circle,fill=white,thick,draw}]
    \node[fill=cyan!50] (\A1) at (\Ox,\Oy) {\fs $V_1$};
    \node[fill=green!50] (\A2) at (8+\Ox,6+\Oy) {\fs $V_2$};
    \node[fill=green!50] (\A3) at (4+\Ox,9+\Oy) {\fs $V_3$};
    \node[fill=red!50] (\A4) at (-4+\Ox,9+\Oy) {\fs $V_4$};
    \node[fill=red!50] (\A5) at (-8+\Ox,6+\Oy) {\fs $V_5$};
    \node[fill=red!50] (\A6) at (-10+\Ox,1.4+\Oy) {\fs $V_6$};
    \node[fill=cyan!50] (\A7) at (-2+\Ox,-9+\Oy) {\fs $V_7$};
    \node[fill=cyan!50] (\A8) at (2+\Ox,-9+\Oy) {\fs $V_8$};
\end{scope}
\begin{scope}[>={Stealth[black]},
              every node/.style={fill=white,rectangle},
              every edge/.style={draw=black,very thick}]
    \path [->] (\A1) edge[bend right=10] node {\fs $W'_{12}$} (\A2);
    \path [->] (\A1) edge[bend left=10] node {\fs $W'_{13}$} (\A3);
    \path [->] (\A1) edge[bend right=10] node {\fs $W'_{14}$} (\A4);
    \path [->] (\A1) edge[bend left=0] node {\fs $W'_{15}$} (\A5);
    \path [->] (\A1) edge[bend left=10] node {\fs $W'_{16}$} (\A6);
    \path [->] (\A1) edge[bend right=10] node {\fs $W'_{17}$} (\A7);
    \path [->] (\A1) edge[bend left=10] node {\fs $W'_{18}$} (\A8);
\end{scope}

\end{tikzpicture}
    \end{center}
    \caption{Illustrating 
    CrossWalk. Different colors represent nodes from different groups.
    Our method upweights edges which are (1) closer to the groups’ peripheries or (2) connecting different groups in the network.
    }
    \label{fig:schema}
\end{figure}

The key idea of our method is to assign larger weights to (1) the edges that are connected to nodes closer to the groups' peripheries, and (2) the edges that are connecting nodes from different groups.
Intuitively, our reweighting method biases 
random walks
initiated in a given group 
towards visiting nodes on the group boundary and eventually 
crossing the boundaries and visiting nodes from other groups in the graph.
A schematic diagram of the proposed method is illustrated in Figure \ref{fig:schema}. 

\subsection{Bias towards Group Boundaries}
For every node $v$ we define a measure of \textit{proximity} to other groups in the graph. 
Intuitively, for every node $v$, its proximity $m(v)$ to other groups indicates the fraction of nodes from other groups in $v$'s close proximity.
To calculate $m(v)$,
we initiate a set of $r$ truncated random walks of length $d$ from a node $v$ based on the original edge weights. We define $v$'s proximity to other groups, $m(v)$, as the expected number of times nodes from other groups are visited in $r$ random walks of length $d$ rooted at $v$. Formally, we have
\begin{equation}\label{eq:proximity}
    m(v)=\frac{\sum_{j\in [r]}\sum_{u\in \mathcal{W}^j_{v}} \mathbb{I}[ l_v \neq l_{u}]}{r \times d}.
\end{equation}
Nodes that are closer to group boundaries and have a larger number of nodes with a different label in their close proximity has a higher value of $m$. 
Assigning larger weights to the edges connecting nodes with larger proximity values biases 
reweighted random walks towards visiting nodes on group boundaries in the graph.

\subsection{Bias towards Other Groups}
In addition, 
to bias the stochastic edge traversal procedure towards visiting nodes from other groups,
we upweight the edges connecting different groups in the graph. More specifically, for a node $v$ we denote 
the set of groups 
in $v$'s immediate neighborhood
by $R_v=\{\cup_{u\in \mathcal{N}(v)} l_u | l_v\neq l_u\}$.
Moreover, we denote the set of $v$'s neighbors within the same group by $N_v=\{u\in \mathcal{N}(v)| l_u=l_v\}$, and the set of $v$'s neighbors that belong to another group $c\neq l_v$ by $N_v^c=\{u|u\in\mathcal{N}(v)| l_v\neq l_u=c\}$.
Now, for parameters 
$\alpha\in(0,1)$,  
$p>0$, we weight $v$'s outlinks 
to its neighbors $u\in \mathcal{N}_v$ as follows: if $u$ belongs to the same group, i.e. $l_u=l_v$, we multiply $w_{vu}$ by $(1-\alpha)\times m(u)^p$, and if $u$ belongs to another group, i.e. $l_v\neq l_u$, we multiply $w_{vu}$ by $\alpha \times m(u)^p$. 
The parameter $p$ controls the degree of biasness of 
random walks towards visiting nodes at group boundaries.
We normalize the edge weights such that the sum of the weights of edges connecting $v$ to its neighbors from the same group is $1-\alpha$, and the sum of the weights of edges connecting $v$ to its neighbors from any other group is $\alpha/|R_v|$.
Formally, the new edge weights can be derived as follows:
\begin{equation}\label{eq:power}
     w'_{vu}\!\!\!\!=\!\!
\begin{cases}
    \!w_{vu}{(1-\alpha)}{}\!\times\!\frac{m(u)^p}{\sum_{z\in N_v}w_{vz}  m(z)^p}, &\!\!\!\! if  u\!\in\!\mathcal{N}(v), l_v=l_u\\
    \!w_{vu}{\alpha}{}\times\!\frac{m(u)^p}{|R_v|\sum_{z\in N_v^c}w_{vz} m(z)^p}, & \!\!\!\!if  u\!\in\!\mathcal{N}(v), l_v\!\neq\! l_u\!\!=\!c.
\end{cases}
\end{equation}

\begin{algorithm}[t] \caption{CrossWalk: Fairness-enhanced node embedding
}
\begin{algorithmic}[1]
\Require Graph $G=(V,E)$, Edge weights $w_{uv}~\forall (v,u)\in E$, Parameters $\alpha, p$.
\Ensure weights $w'_{vu} ~\forall (v,u)\in E$.
\For{$v \in V$} \Comment{Calculating closeness to boundary}
\State Run $r$ random walks $\mathcal{W}_v^j, j\in [r]$ rooted at $v$.
\State $m(v)={\sum_{j\in [r]}\sum_{u\in \mathcal{W}^j_{v}} \mathbb{I}[ l_v \neq l_{u}]}/{(r \times d)}.$
\EndFor
\For{$v \in V$} \Comment{Reweighting edges}
\State $N_v=\{u\in \mathcal{N}(v)| l_u=l_v\}$
\State $N_v^c=\{u\in \mathcal{N}(v)| l_v\neq l_u=c\}$
\State $R_v=\{\cup_{u\in \mathcal{N}(v)} l_u | l_v\neq l_u\}$
\State $Z=\sum_{u\in N_v}w_{vu} \times m(u)^p$
\For{$u \in N_v$}
\Comment{Edges in same group}
\State $w^{'}_{vu}=w_{vu} \times {(1-\alpha)} \times {m(u)^p}/{Z}$
\EndFor
\For{$c\in R_v$}
\State $Z=|R_v| \times \sum_{u\in N_v^c} w_{vu} \times m(u)^p$
\For{$u \in N_v^c$} \Comment{Edges connecting different groups}
\If{$N_v\neq\emptyset$}
\State $w^{'}_{vu}=w_{vu} \times {\alpha}{} \times {m(u)^p}/{Z}$
\Else
\State{ $w^{'}_{vu}=w_{vu} \times  {m(u)^p}/{Z}$}
\EndIf
\EndFor
\EndFor
\EndFor
\end{algorithmic}\label{alg}
\end{algorithm}
Larger $\alpha$ upweights edges connecting different groups, and biases the stochastic edge traversal procedure towards visiting nodes belonging to other groups in the graph.
%
The pseudocode of CrossWalk is shown in Alg.\ref{alg}.

\subsection{CrossWalk Enhances Fairness}\label{sec:intuition}
\begin{figure*}[!htb]
    \begin{center}
    \subfloat[Synthetic Layered Dataset \\ DeepWalk \label{fig:tsne-synthetic3layers-unwighted}]{
        \includegraphics[width=0.24\textwidth]{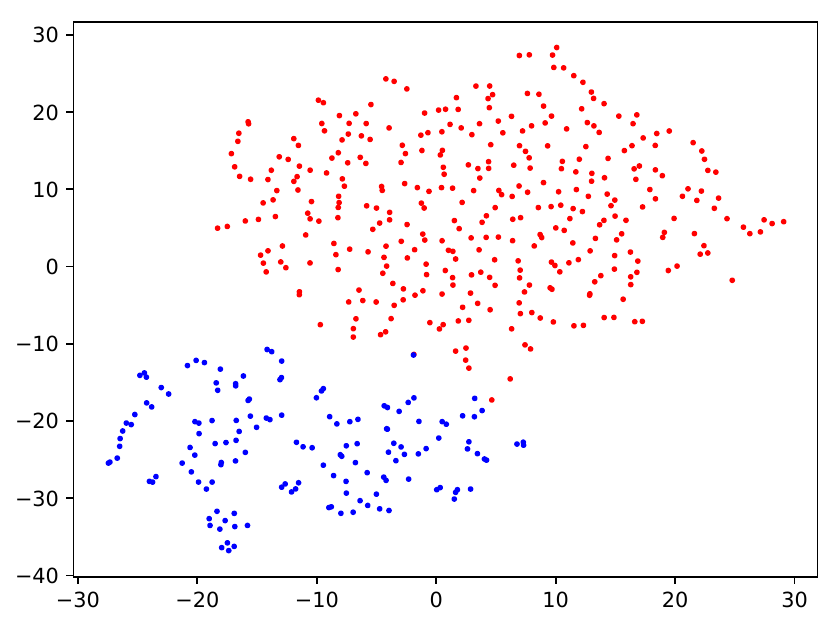}
    }
    \subfloat[Synthetic Layered Dataset \\ CrossWalk ($\alpha=0.5$, $p=2$) \label{fig:tsne-synthetic3layers-reweighted}]{
        \includegraphics[width=0.24\textwidth]{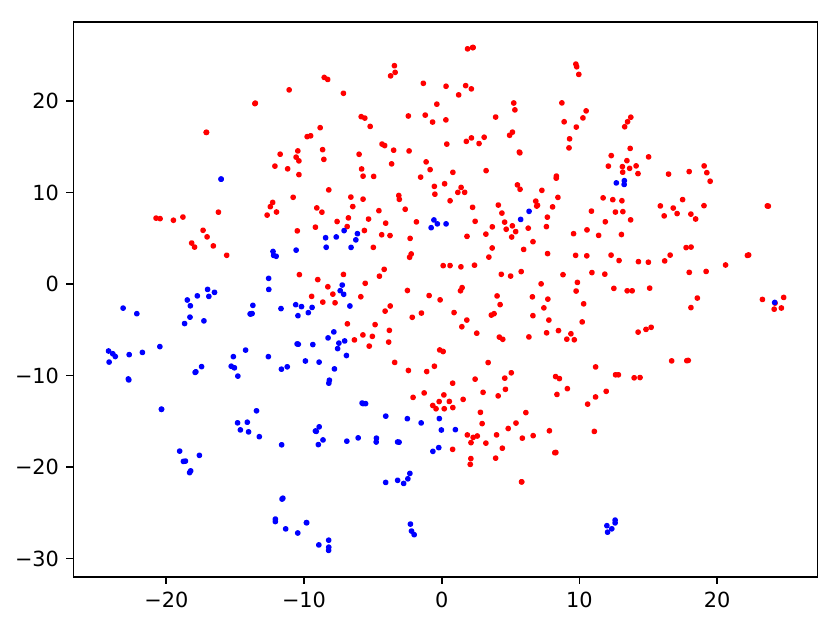}
    }
    \subfloat[Rice-Facebook Dataset \\ DeepWalk \label{fig:tsne-rice-unwighted}]{
        \includegraphics[width=0.24\textwidth]{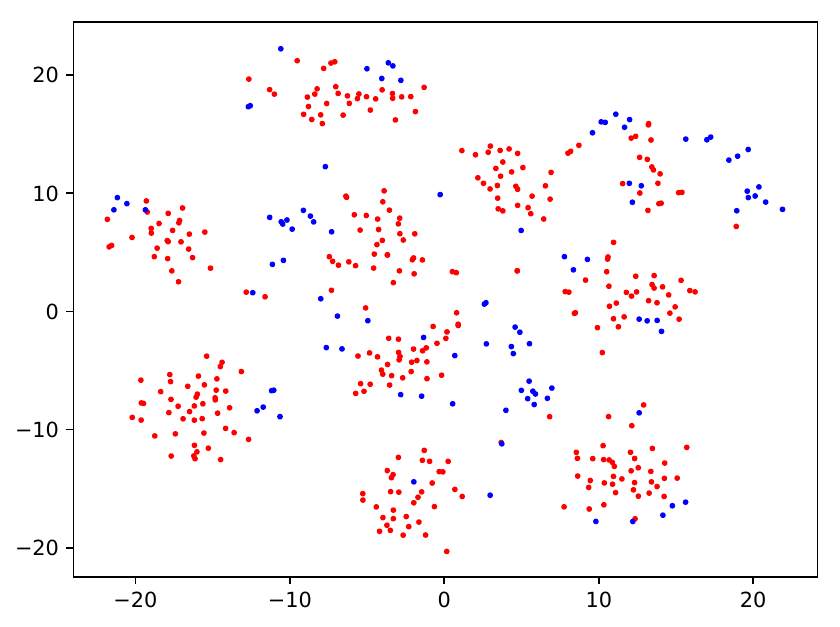}
    }
    \subfloat[Rice-Facebook Dataset \\ CrossWalk ($\alpha=0.5$, $p=4$) \label{fig:tsne-rice-reweighted}]{
        \includegraphics[width=0.24\textwidth]{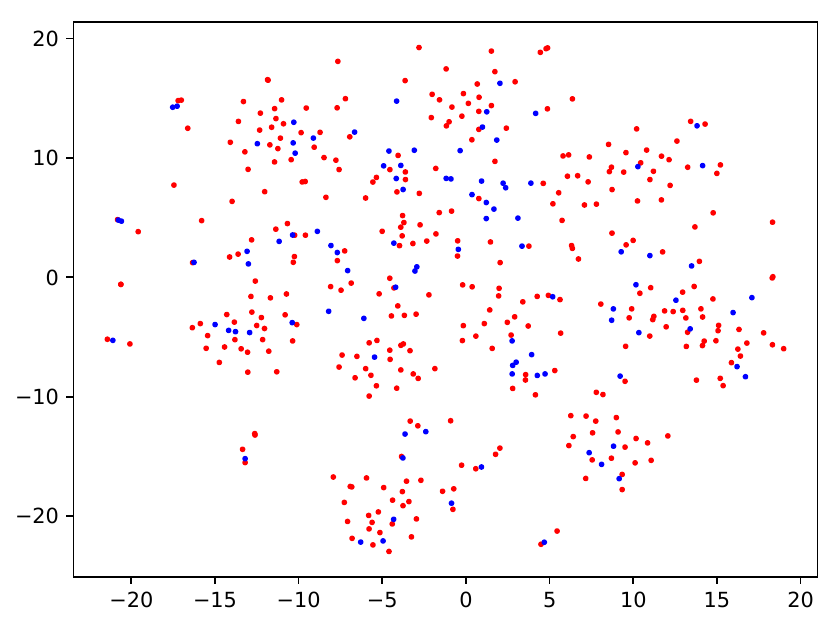}
    }
    \end{center}
    \caption{Distributions of the {DeepWalk} embedded nodes from the two groups. Embeddings of the two groups are pulled towards each other in the graph reweighted by CrossWalk.}
    \label{fig:tsne}
\end{figure*}

Our reweighting method upweights the edges that are closer to the groups' peripheries and those connecting different groups.
When transition probabilities in random walks are chosen based on our reweighting strategy, the random walks initiated by representation learning algorithms spend more time visiting nodes at group boundaries.
Hence, CrossWalk pulls nodes that are near groups’ peripheries towards their neighbors from other groups in the embedding space, while it preserves the necessary structural properties of the graph, because just reweights the existing edges and does not allow normal random walks to randomly skip over nodes.
Figure \ref{fig:tsne} compares the {DeepWalk} projection of node embeddings of the original and the reweighted graph by CrossWalk on a 2-D space.
We observe that the embeddings of the two groups are pulled towards each other in the reweighted graph. 
For networks in which the majority of nodes are more than one hop away from group peripheries, larger value of $p$ bias random walks towards other groups and result in fairness-enhanced representations. Moreover, for networks with more inter-group connections, larger values of $\alpha$ should be used.
We experimentally study the effect of $\alpha, p$ in Section \ref{sec:appendix-parameter-tuning} of Supplementary Material.

\begin{figure}[t]
  \centering
    \scalebox{0.35}{
\def \sc {0.3}
\begin{tikzpicture}[scale=\sc]

\def \Ox {0}
\def \Oy {0}
\def \fs {\huge}

\node (red) at (12+\Ox,-1.5+\Oy) [draw,red,ellipse,very thick,dashed,minimum width=\sc*30cm,minimum height=\sc*18cm] {};

\begin{scope}[every node/.style={circle,fill=red!50,thick,draw, minimum size=25, font=\fs}]
    \node (R1) at (0+\Ox,0+\Oy) {$w$};
    \node (R2) at (10+\Ox,0+\Oy) {};
    \node (R3) at (16+\Ox,-2+\Oy) {};
    \node (R4) at (15+\Ox,5+\Oy) {};
    \node (R5) at (11+\Ox,-8+\Oy) {};
    \node (R6) at (23+\Ox,0+\Oy) {$u$};
\end{scope}

\begin{scope}[every node/.style={fill=white,font=\fs}]
    \node (S) at (6+\Ox,4+\Oy) {$S$};
\end{scope}

\begin{scope}[>={Stealth[black]},
              every edge/.style={draw=black,very thick}]
\path [-] (R2) edge node {} (R4);
\path [->] (R2) edge node {} (R3);
\path [->] (R3) edge node {} (R6);
\path [-] (R4) edge node {} (R5);
\path [-] (R4) edge node {} (R6);
\path [-] (R5) edge node {} (R6);
\end{scope}

\begin{scope}[>={Stealth[black]},
              every edge/.style={draw=black,very thick, dashed}]
\path [->] (R1) edge node {} (R2);
\end{scope}

\node (blue) at (40+\Ox,9.2+\Oy) [draw,blue,ellipse,very thick,dashed,minimum width=\sc*22cm,minimum height=\sc*11cm, rotate=-15] {};

\begin{scope}[every node/.style={circle,fill=cyan!50,thick,draw, minimum size=25}]
    \node (B1) at (35+\Ox,8+\Oy) {};
    \node (B2) at (43+\Ox,6+\Oy) {};
    \node (B3) at (35+\Ox,13+\Oy) {};
    \node (B4) at (43+\Ox,11+\Oy) {};
\end{scope}

\begin{scope}[every node/.style={circle,fill=cyan,thick,draw, minimum size=1,inner sep=1.5pt}]
    \node (B5) at (46+\Ox,8.5+\Oy) {};
    \node (B6) at (47.5+\Ox,8+\Oy) {};
    \node (B7) at (49+\Ox,7.5+\Oy) {};
\end{scope}

\begin{scope}[>={Stealth[black]},
              every edge/.style={draw=black,very thick}]
\path [-] (B1) edge node {} (B2);
\path [-] (B3) edge node {} (B4);
\path [-] (B1) edge node {} (B4);
\path [->] (R6) edge node {} (B1);
\path [-] (R6) edge node {} (B3);
\end{scope}

\node (green) at (40+\Ox,-9.2+\Oy) [draw,green,ellipse,very thick,dashed,minimum width=\sc*22cm,minimum height=\sc*10cm, rotate=15] {};

\begin{scope}[every node/.style={circle,fill=green!50,thick,draw, minimum size=25}]
    \node (G1) at (35+\Ox,-8+\Oy) {};
    \node (G2) at (40+\Ox,-6+\Oy) {};
    \node (G3) at (35+\Ox,-13+\Oy) {};
    \node (G4) at (43+\Ox,-11+\Oy) {};
\end{scope}

\begin{scope}[every node/.style={circle,fill=green,thick,draw, minimum size=1,inner sep=1.5pt}]
    \node (G5) at (46+\Ox,-8.5+\Oy) {};
    \node (G6) at (47.5+\Ox,-8+\Oy) {};
    \node (G7) at (49+\Ox,-7.5+\Oy) {};
\end{scope}

\begin{scope}[>={Stealth[black]},
              every edge/.style={draw=black,very thick}]
\path [-] (G1) edge node {} (G2);
\path [-] (G1) edge node {} (G3);
\path [-] (G2) edge node {} (G4);
\path [-] (R6) edge node {} (G1);
\path [-] (G2) edge node {} (B1);
\path [-] (G4) edge node {} (B2);
\end{scope}

\end{tikzpicture}
}
    \caption{{\footnotesize Example illustrating the importance of  the proximity metric utilized in CrossWalk. It biases random walks initiated from internal nodes of $S$, such as $w$, towards passing $u$ and visiting other groups in few steps.}}
    \label{fig:example}
\end{figure}
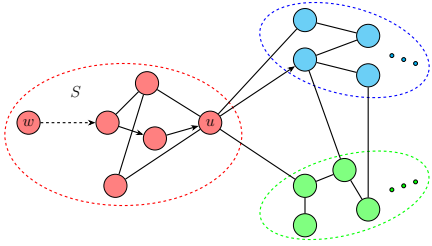


\subsection{CrossWalk vs FairWalk}
Here, we discuss how our method leverages the proximity measure $m(v)$ in Eq. \eqref{eq:proximity} to overcome the shortcoming of existing methods such as FairWalk \cite{rahman2019fairwalk}, and achieve a superior performance.
In a graph with $C$ groups, FairWalk only 
reweights the edges between different groups, such that all the groups $c'\in [C]$ have equal probability $1/C$ of being visited from a boundary node in group $c\in [C]$.
Crucially, there are two major differences between CrossWalk and FairWalk. First,
CrossWalk carefully reweights inner-group connections, by assigning larger weights to edges that are closer to group peripheries. 
In doing so, random walks started within a group will be biased towards the group's peripheries.
Using a larger power $p$ for the proximity $m(v)^p$ in Eq. \eqref{eq:power} further increases the bias of the random walks towards group boundaries.
In contrast, FairWalk does not bias random walks initiated within a group towards visiting nodes at the group's peripheries.
Hence, FairWalk fails to enhance fairness in graphs where the majority of nodes are more than one hop away from group peripheries.
%
Second, 
CrossWalk has a tunable parameter $\alpha$ to reweight the edges between different groups, 
such that all the groups $c'\in [C]$, for $c'\neq c$ have equal probability $\alpha/(C-1)$ of being visited from a boundary node in group $c\in [C]$.
Hence, CrossWalk always considers a constant probability $1 - \alpha$ for a random walk to stay within the same group. The use of $\alpha$, in addition to $p$, allows adjusting the amount that node representations from different groups are pulled towards each others.
Moreover, $\alpha$ becomes important in extreme situations where a boundary node in a minority group $c\in C$ is connected to several other groups $c'\in C$, for $c\neq c'$. In such cases, FairWalk has a small probability of staying within the same group $c$, while CrossWalk stays within the same group with a constant probability of $1 - \alpha$. 
Larger values for $p$ and $\alpha$ pull the embedding of the nodes in different groups further towards each other, and reduce the disparity of graph algorithms applied to the embeddings. An experimental study of the effect  of $\alpha$ and $p$ can be found in Section \ref{sec:appendix-parameter-tuning} of Supplementary Material.

To illustrate the importance of our proximity metric (see Figure \ref{fig:example}), consider a small group of nodes $S$, where all paths to other groups go through node $u \in S$ i.e., $u$ is the only gateway of $S$. Indeed, here $u$ takes the largest value of proximity, and nodes in $S$ close to $v$ take high proximity measures. Suppose a random walk starts from a node $w \in S$, $w \neq u$. The {proximity} metric in CrossWalk biases the random walks towards visiting $u$ and meeting other groups in a few steps. This pulls the final embedding of the nodes in $S$ towards those of the other groups, and reduces the disparity of graph algorithms on the embeddings. In contrast, with FairWalk, a random walk starting from $w$ is not biased to visit $u$ soon. In this case FairWalk does not differ from a normal random walk on the original graph, until it visits $u$.

\section{Applications} \label{sec:app-link-pred}
In this section, we discuss applications of our proposed reweighting strategy to enhance fairness 
in the result of random walk based graph algorithms, 
including 
influence maximization, node classification, and link predictions in graphs.  
In Section \ref{sec:appendix-classical-IM} of Supplementary Material, we consider the classical influence maximization problem based on the Independent Cascade (IC) model, and discuss the application of our reweighting method to reduce disparity of influenced individuals from various groups. 

\vspace{2mm}\noindent\textbf{Influence Maximization}
Having the node representations learned by CrossWalk applied to DeepWalk or Node2Vec, the most influential individuals can be found as the set of most centrally located nodes in the representation space. These nodes are medoids of the node representations, and can be found by minimizing the pairwise dissimilarities between nodes within a cluster and a node designated as the center of that cluster in the representation space. For a specific value of $k$, the set of $k$-medoids can be found as follows:
\begin{align}\label{eq:medoid}
    S^* \in & {\arg\min_{\substack{S\subseteq V,\\|S|\leq k}}}\sum_{i \in V} \min_{j \in S}
    \|\Phi_i-\Phi_j\|_2  
\end{align}
A common algorithm to find the set of $k$-medoids starts by selecting $k$ nodes uniformly at random. Then, it iteratively assigns each node to the cluster defined by the nearest medoid, updates the medoids within the new clusters, and repeats this procedure as long as the sum of distances between the nodes and their corresponding medoid in the representation space is decreasing.

We calculate the number of infected individuals by simulating diffusion started from the selected seeds $S^*$ in the {original} graph that is not reweighted by CrossWalk. We use Independent Cascade (IC) model, in which
at every time step $t>0$, a node $u \in V$ which was activated at time $t-1$ can activate its inactivated neighbor $v$ with probability $P_{uv}$. 
The diffusion stops at time $t>0$ if no new node gets activated.  
In a weighted network, a multiplication of the original edge weights $w_{uv}$ can be used as transmission probabilities $P_{uv}$ in the IC model.

\vspace{2mm}\noindent\textbf{Node Classification}
Node representations learned by CrossWalk applied to DeepWalk and Node2Vec can be used for node classification in graphs. 
Here, we consider Label Propagation (LP) 
to classify nodes based on the obtained representations.
LP first constructs a $k$-nearest graph from the data and initializes all the nodes with a unique label. Then, it iteratively assigns to every node the label with the highest frequency among its neighbors. This process is repeated until every node has the same label as the majority of its neighbors in the graph.

\vspace{2mm}\noindent\textbf{Link Prediction}
Node representations obtained by CrossWalk applied to DeepWalk and Node2Vec can be  used 
to detect new or formerly unknown connections in a network. 
To do so, we train a logistic regression on edges. 
For nodes $v$ and $u$ with representation vectors $r_v$ and $r_u$, 
the feature vector used for the link $(v,u)$ is $(r_v \!-\! r_u)^{\circ 2}$, where $\circ$ denotes Hadamard power. For evaluation, we randomly select 10\% of the existing edges as positive test data and do not use them when obtaining node embeddings. We also select equal number of non-existing edges as negative test data. The rest of the existing edges in the graph and an equal number of non-existing edges are used for training.

\vspace{2mm}\noindent\textbf{CrossWalk \& Graph Algorithms}
As discussed in Sec. \ref{sec:intuition}, CrossWalk pulls together representation of nodes belonging to different groups.
In the influence maximization problem, the most centrally located nodes in the representation space will have a larger proximity to multiple groups. Therefore, the selected seeds can spread the information more effectively in multiple groups.
Similarly, in the node classification and edge prediction problems, as similar nodes or edges belonging to various groups become closer in the representation space, they share the same label irrespective of their group. This enhances fairness and decreases the disparity of test accuracy in different groups.

\section{Experiments} \label{sec:experiments}
In this section we evaluate the effectiveness of CrossWalk for reducing disparity of influence maximization, link prediction and node classification on two real and two synthetic datasets.  
We use DeepWalk \cite{deepwalk} and FairWalk \cite{rahman2019fairwalk} as baseline methods. For influence maximization, we also use the greedy algorithm and Adversarial graph embedding \cite{khajehnejad2020adversarial} as baseline methods. In all experiments CrossWalk and FairWalk use DeepWalk embedding, unless we explicitly mention that Node2vec is used. Our experiments confirm the superiority of CrossWalk over baselines in different tasks.
Source code and datasets used in our experiments are available anonymously from \url{https://github.com/ahmadkhajehnejad/CrossWalk}.  


\vspace{2mm}\noindent\textbf{Rice-Facebook Dataset:}
This dataset \cite{mislove2010you} is an undirectd graph of friendships between students at the Rice university. The dataset contains 1205 nodes and 42443 edges. Node features include age, college and major.
We use students' ages as the sensitive attribute. We consider the students with age 20 as group A and the students with ages 18 and 19 as group B, and exclude the nodes with ages higher than 20. Group A has 344 nodes and 7441 inner-group connections. Group B has 97 nodes and 513 inner-group connections. There are 1779 connections between the two groups.

\vspace{2mm}\noindent\textbf{Twitter Dataset:}
We consider an undirected and connected sub-graph of the Twitter dataset \cite{babaei2016efficiency,cha_icwsm10} with 3560 nodes. The nodes can be divided into three groups based on their political learning: neutrals (group A) with $2598$ nodes, (group B) liberals with $782$ nodes, and (group C) conservatives with $180$ nodes. The number of intra-group and inter-group connections are $e_{intra}^A =3724$, $e_{intra}^B =950$, $e_{intra}^C =74$, $e_{inter}^{AB} = 1461$, $e_{inter}^{AC} = 359$ and $e_{inter}^{BC} = 109$. 

\vspace{2mm}\noindent\textbf{Synthetic Datasets:}
We consider two undirected synthetic datasets. Our first synthetic network consists of two groups with $n_A\!=\!350$ and $n_B\!=\!150$ nodes that are connected with intra-group probabilities $P_{intra}^A\!=\!P_{intra}^B- 0.025$ and inter-group probability $P_{intere}^{AB}\!=\!0.001$. 

Our second synthetic network consists of three groups with $n_A=300$, $n_B=125$ anad $n_C=75$ nodes. Nodes are connected with intra-group probabilities of $P_{intra}^A = P_{intra}^B = P_{intra}^C = 0.025$ and inter-group probabilities of $P_{inter}^{AB}=0.001$ and $P_{inter}^{AC} = P_{inter}^{BC} = 0.0005$.

\newcommand{\w}{.23}

\begin{figure}[t]
    \begin{center}
    \subfloat[Rice-Facebook,\\ 
    \centerline{$\alpha=0.5$, $p=4, k=40$.}
    \label{fig:rice}]{
        \includegraphics[width=0.23\textwidth]{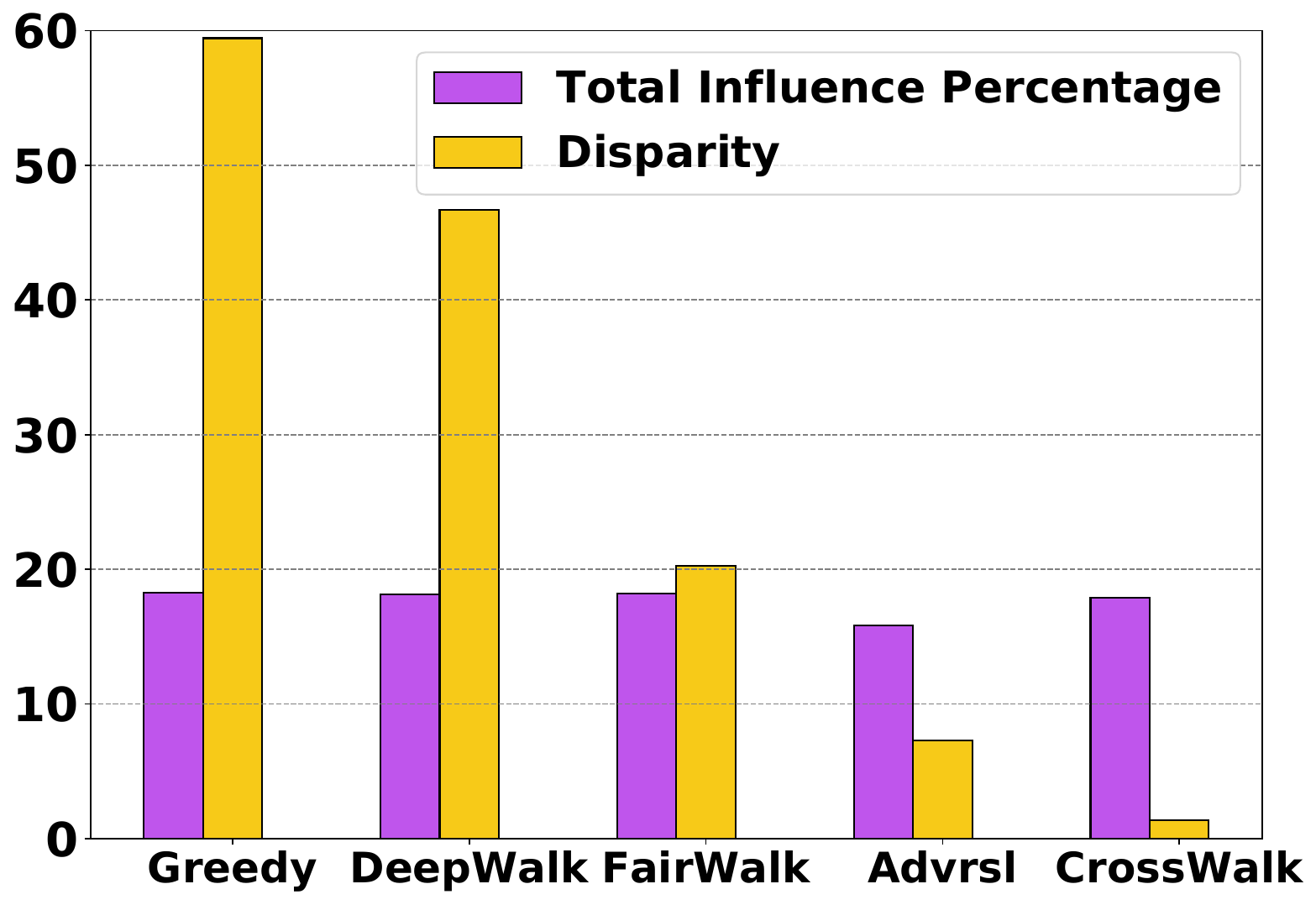}
    }
    \subfloat[2-grouped synthetic\\ 
    \centerline{dataset, $\alpha=0.7$, $p=4$.}
    \label{fig:synth2}]{
        \includegraphics[width=0.23\textwidth]{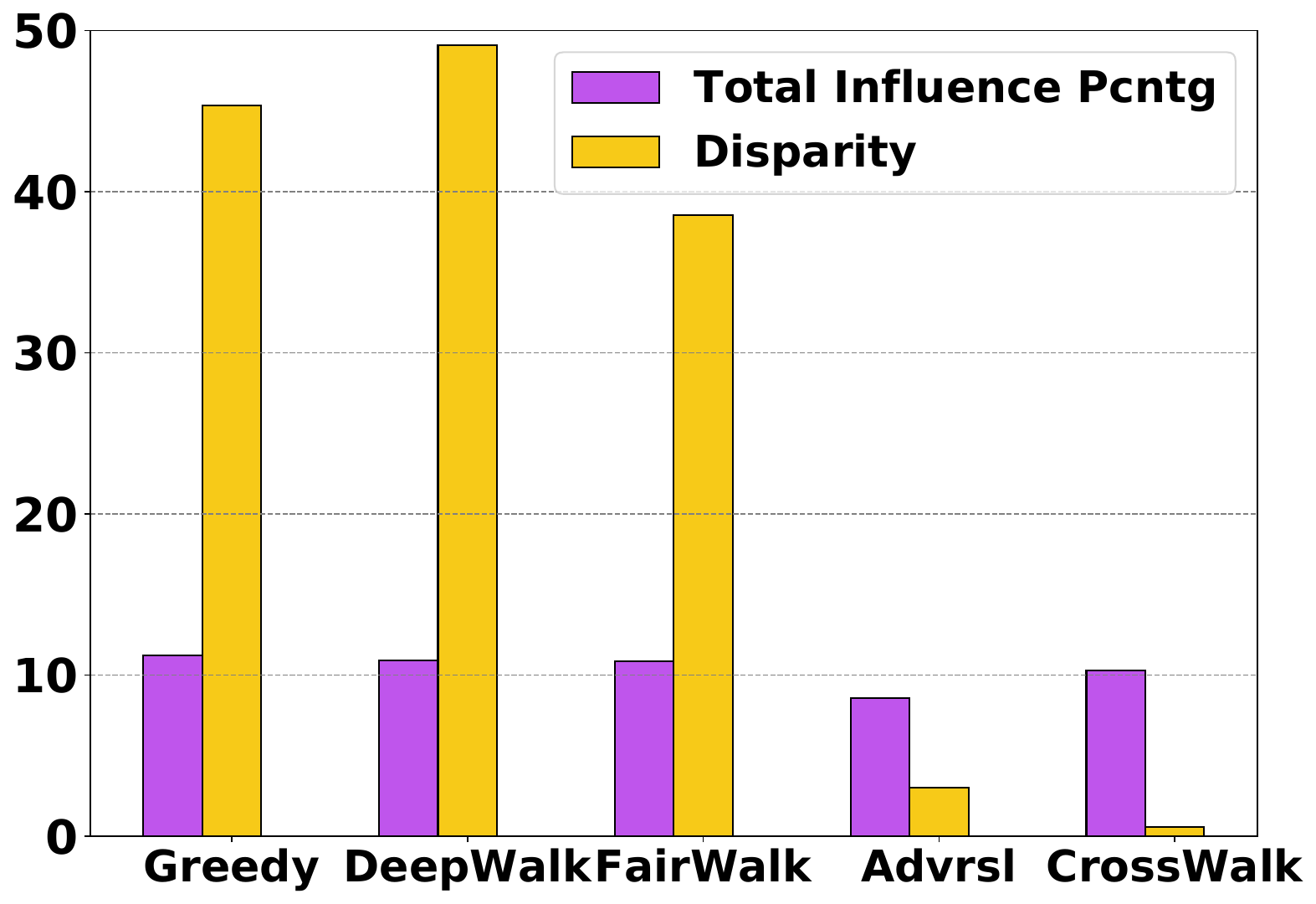}
    }
    \\
    \subfloat[Twitter dataset,\\
    \centerline{$\alpha=0.5$, $p=2$.}
    \label{fig:sample_4000_connected_subset}]{
        \includegraphics[width=0.24\textwidth]{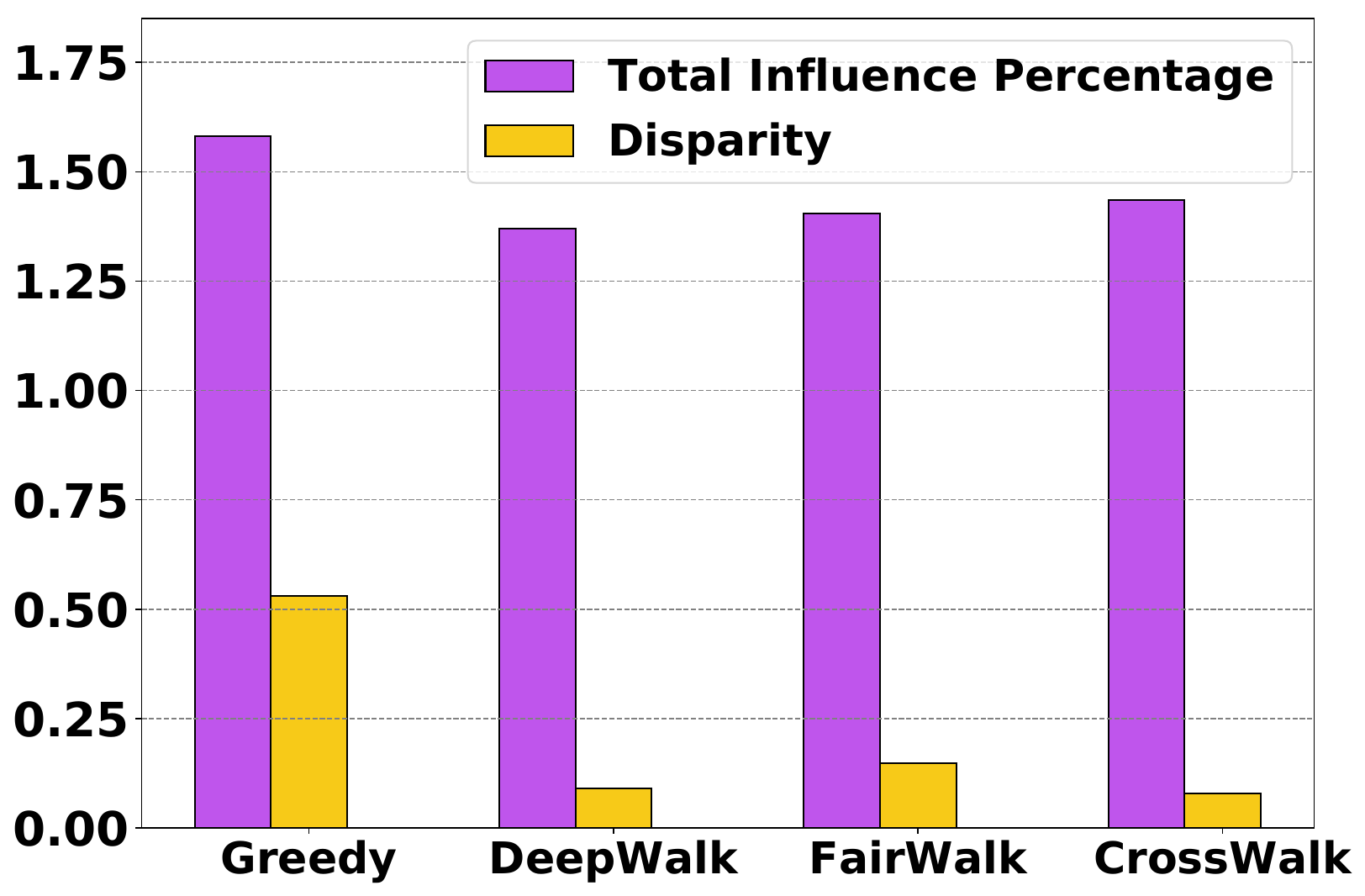}
    }
    \subfloat[3-grouped synthetic\\
    \centerline{dataset, $\alpha=0.7$, $p=4$.}
    \label{fig:synth3}]{
        \includegraphics[width=0.23\textwidth]{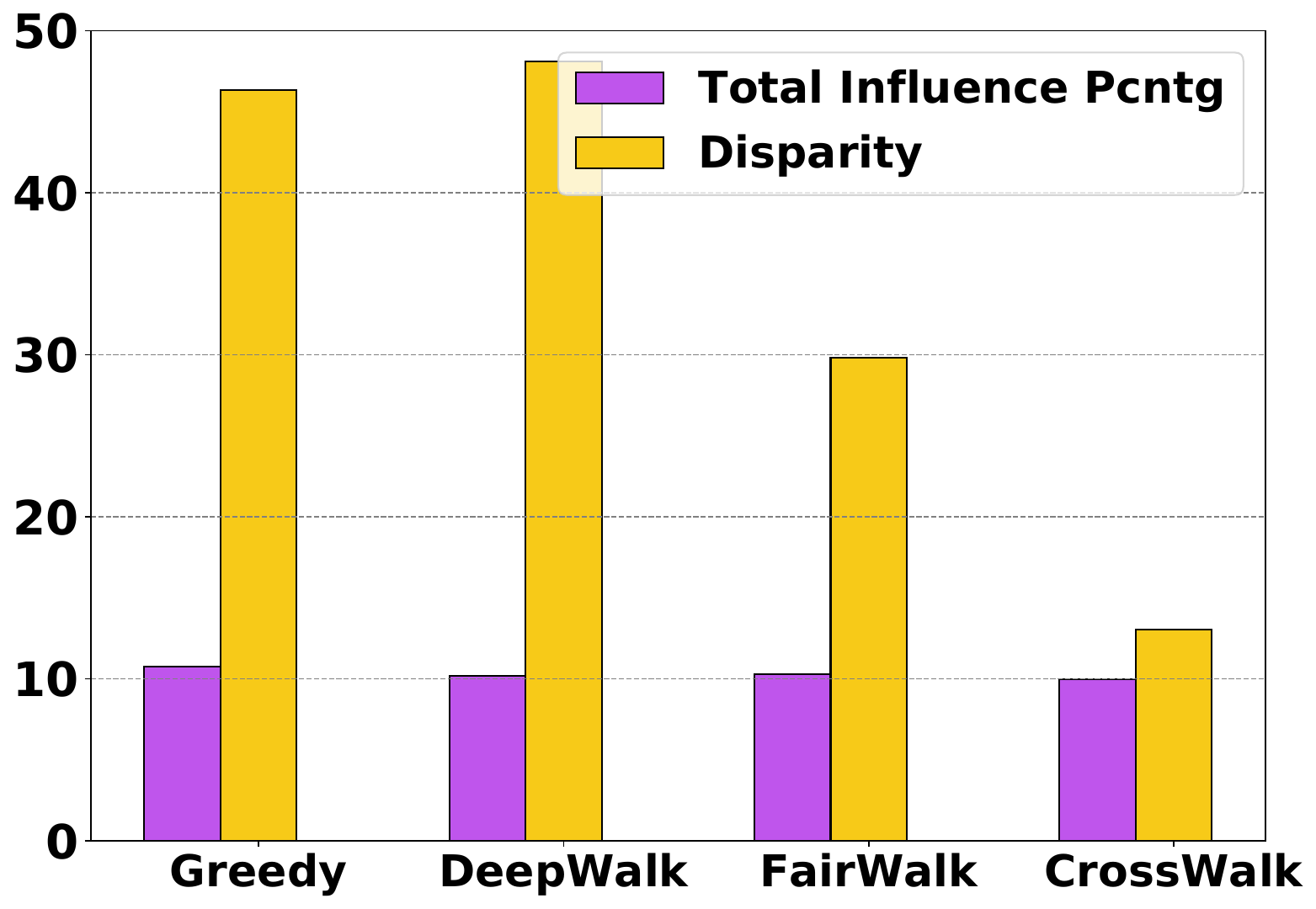}
    }
    \end{center}
    \vspace{-5mm}
    \caption{Influence Maximization}
    \vspace{3mm}
    \label{fig:IM}
\end{figure}

\begin{figure}[!tbp]
  \centering
  \begin{minipage}[b]{0.23\textwidth}
    \includegraphics[width=\textwidth]{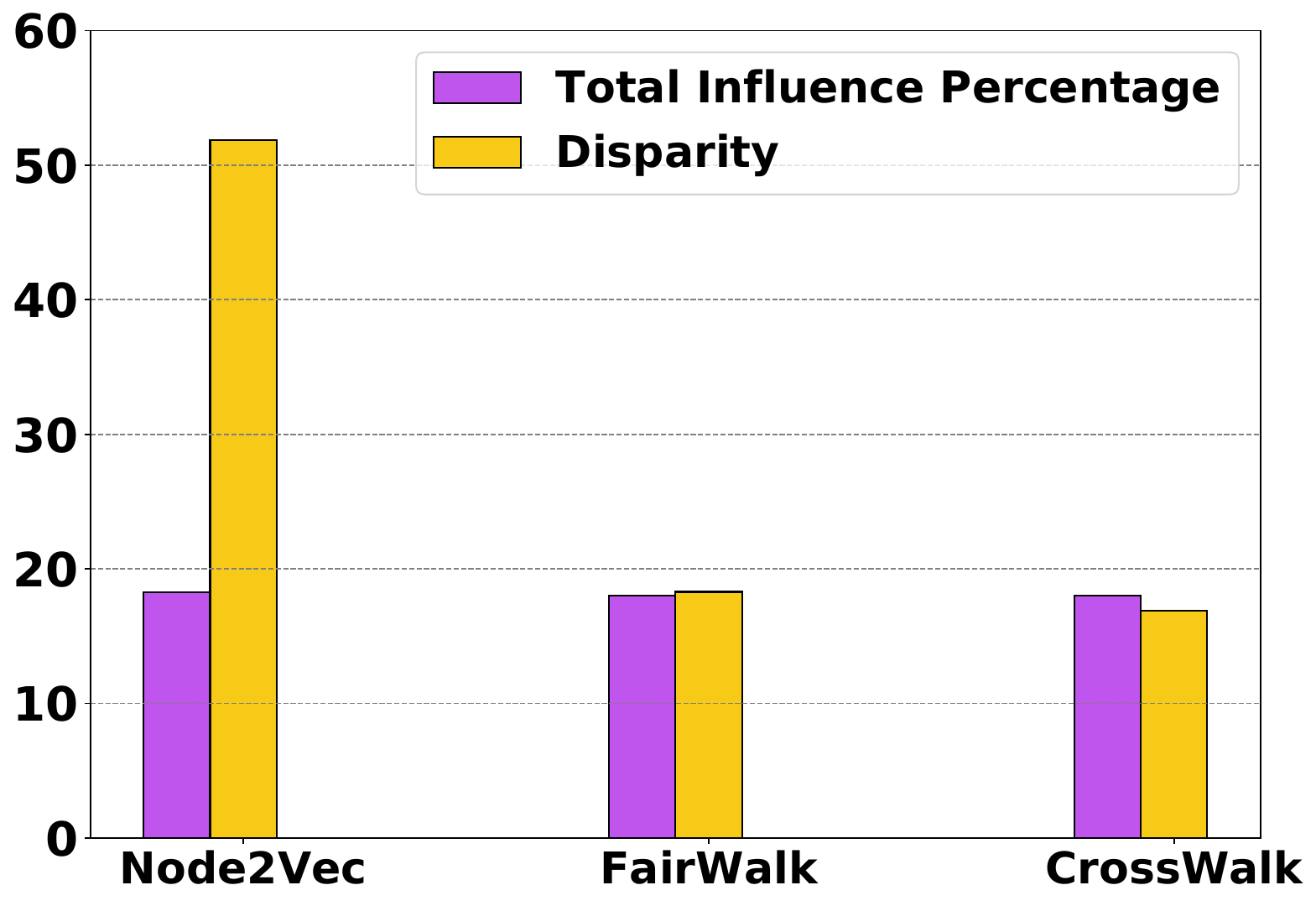}
    \caption{{\footnotesize Influence Maximization - CrossWalk and FairWalk on Node2vec }}
    \label{fig:node2vec}
  \end{minipage}
  \hfill
  \begin{minipage}[b]{0.235\textwidth}
    \includegraphics[width=\textwidth]{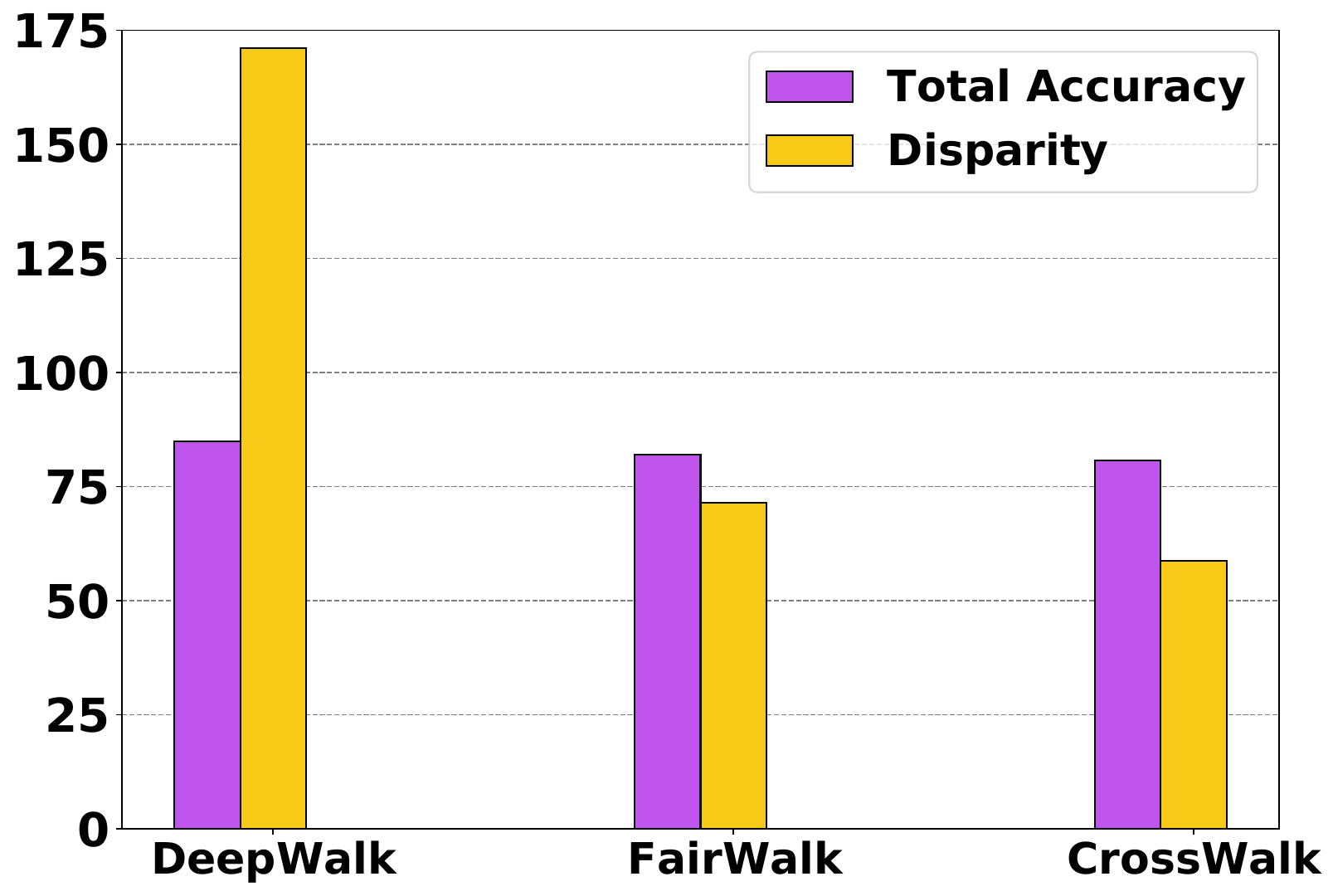}
    \caption{{\footnotesize Node Classification - Rice-Facebook dataset}.
  \footnotesize $\alpha=0.5$, $p=1$.}
  \vspace{0.07cm}
    \label{fig:classification-rice-LP}
  \end{minipage}
\end{figure}

\subsection{Influence Maximization} \label{sec:results-infmax}
To find the set of most influential nodes, we apply $k$-medoids with $k=40$ to node representations (by different methods) to select the seeds. Moreover, we run the baseline greedy seed selection algorithm to show that in most cases, the total influences obtained by embedding based seed selection methods are comparable with the well known greedy seed selection algorithm.
We also report the performance of Adversarial Embedding \cite{khajehnejad2020adversarial} as a baseline for fair influence maximization. This method is only applicable to networks of 2 groups.
To calculate the number of infected individuals, we consider Independent Cascade model (IC) with a constant activation probability for all the edges in the original network (0.01 for real datasets and 0.03 for synthetic networks).
We calculate disparity according to Eq. \eqref{eq:fairness-def},
and report the averaged results of 5 runs for each embedding based method. 

Figures \ref{fig:rice} and \ref{fig:synth2} show total influence and disparity of different methods on the Rice-Facebook dataset and the 2-grouped synthetic dataset, respectively. Comparing with DeepWal, we observe that CrossWalk results in a much larger decrease in disparity than FairWalk and Adversarial embedding, with a very small decrease in the total influence.
Figures \ref{fig:sample_4000_connected_subset} and \ref{fig:synth3} demonstrate the same results for the Twitter and our second synthetic dataset, each with 3 groups. Interestingly we see on Twitter dataset that FairWalk increases the disparity compared to DeepWalk, but CrossWalk both decreases the disparity and improves the total influence.

We also apply CrossWalk to Node2Vec on Rice-Facebook dataset. Figure \ref{fig:node2vec} compares the performance of CrossWalk with $\alpha=0.5, p=4$, and FairWalk applied to Node2Vec with $p=0.5, q=0.5$. 
We observe that CrossWalk outperforms FairWalk in reducing the disparity.

\subsection{Node Classification}
We use the Rice-Facebook dataset for classification, using the students' ages as the sensitive attribute and their college IDs as their class labels. We randomly partition the nodes into two equal sized training and test sets.
We measure disparity according to Equation \eqref{eq:fairness-def}.
Figure \ref{fig:classification-rice-LP} shows the accuracy and disparity of Label Propagation (LP) with $k=7$, applied to the representations obtained by FairWalk and CrossWalk applied to DeepWalk. We report the average result over 200 runs. We see the superiority of CrossWalk in reducing the disparity of node classification. 

\begin{figure}[t]
    \begin{center}
    \subfloat[Rice-Facebook Dataset,\\
    \centerline{$\alpha=0.5$, $p=2$.}
    \label{fig:link-prediction-rice}]{
        \includegraphics[width=0.23\textwidth]{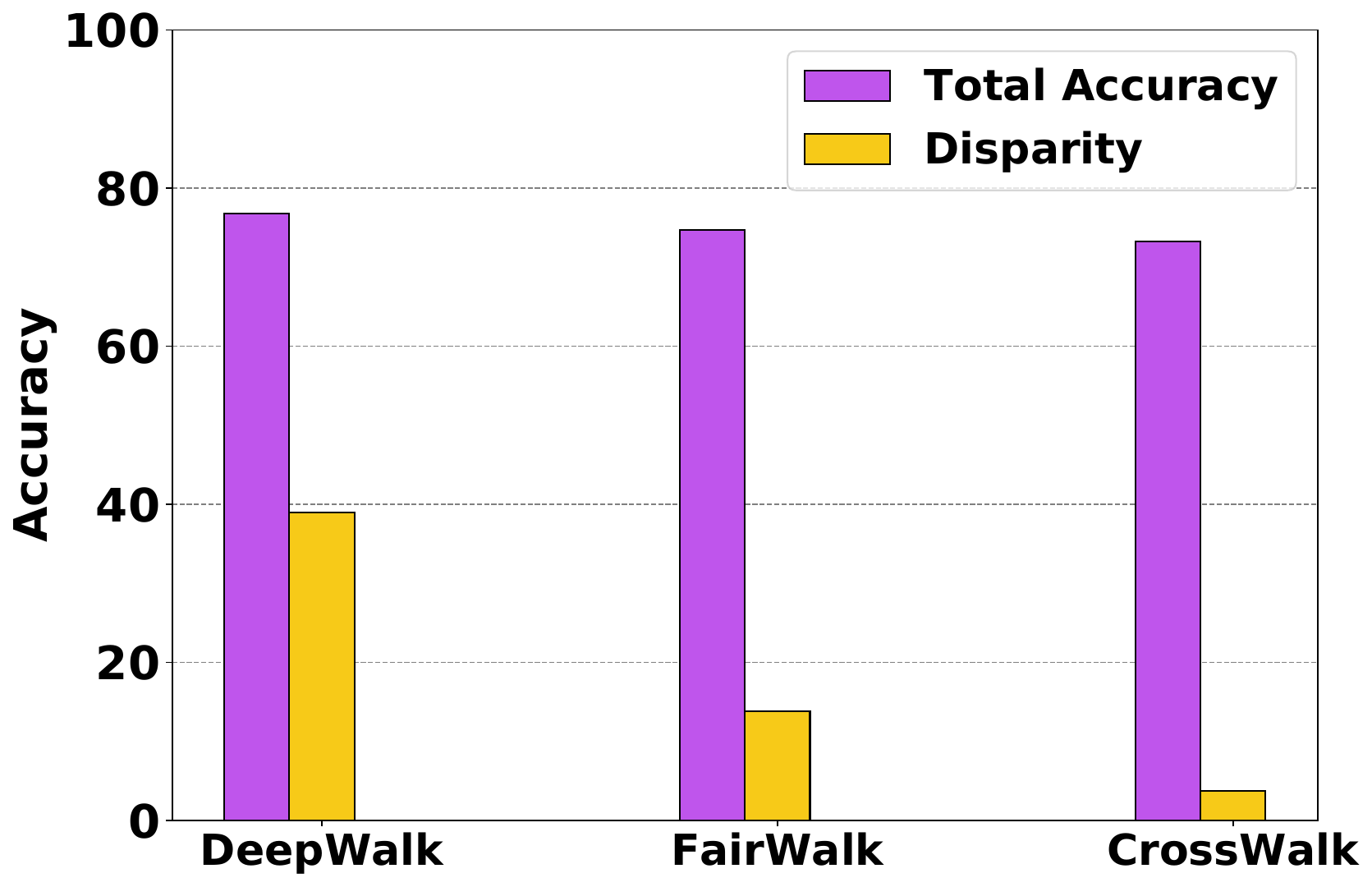}
    }
    \subfloat[Twitter Dataset,\\
    \centerline{$\alpha=0.5$, $p=2$}
    \label{fig:link-prediction-sample-4000-connected-subset}]{
        \includegraphics[width=0.23\textwidth]{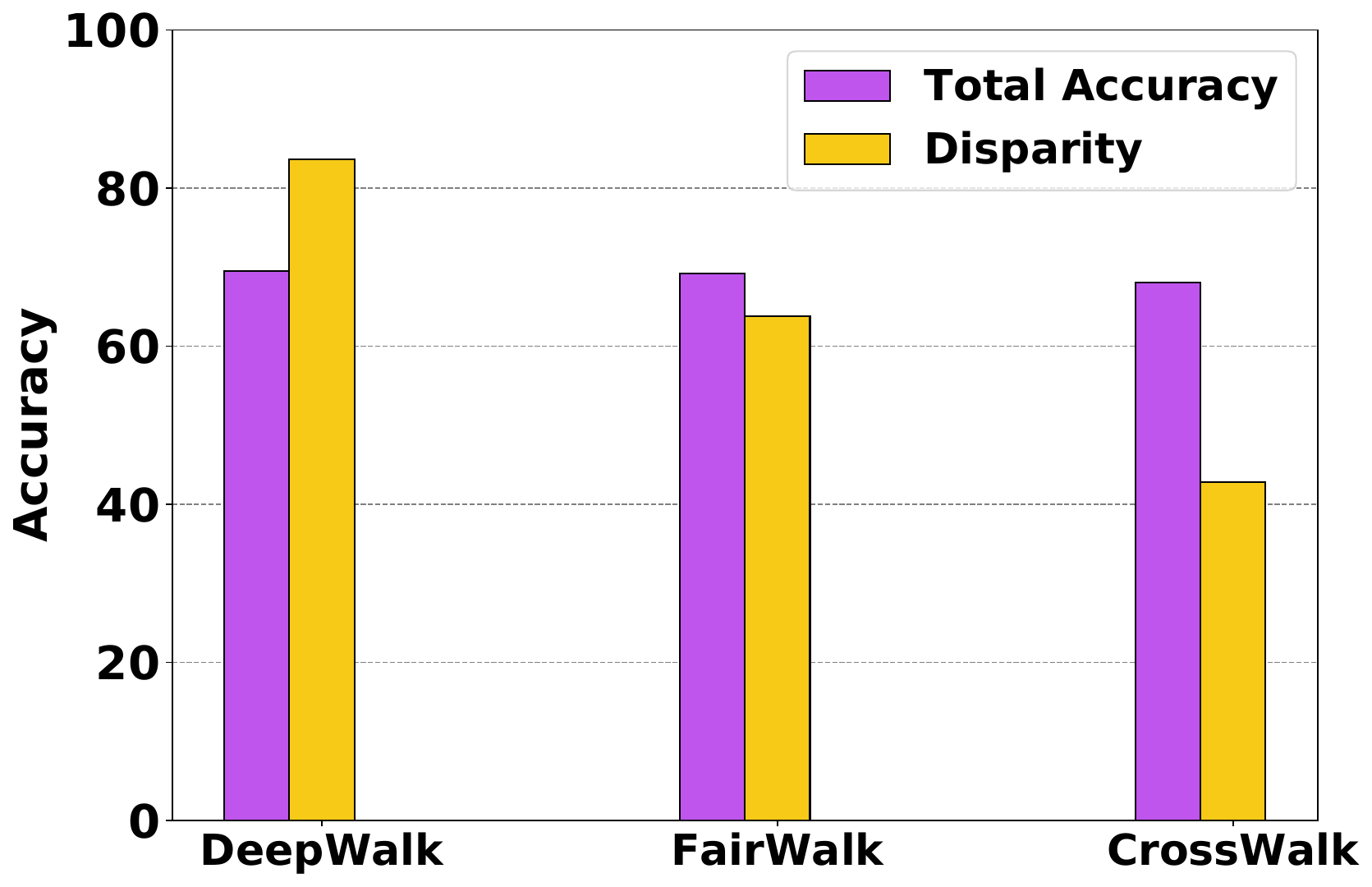}
    }
    \end{center}
    \vspace{-5mm}
    \caption{Link Prediction}
    \label{fig:link-prediction}
\end{figure}

\subsection{Link Prediction}
In the Rice-Facebook dataset with two groups of nodes, A and B, there exist three types of links; A to A, B to B, and A to B connections. Similarly, in the Twitter dataset with three groups of nodes there exist six types of links. For each connection type, we select equal number of positive and negative test samples (10\% of each group).
We train a logistic regression on the embeddings obtained by FairWalk and CrossWalk applied to DeepWalk (see Section \ref{sec:app-link-pred} for more details).
We report the average result over 5 runs. 
Figures \ref{fig:link-prediction-rice} and \ref{fig:link-prediction-sample-4000-connected-subset} illustrate the total accuracy and disparity according to Eq. \eqref{eq:fairness-def}, for link prediction. 
It confirms the superiority of CrossWalk over FairWalk in reducing disparity
with a slight decrease in accuracy. 

\section{Conclusion}
In this work, we developed a simple, and effective approach to enhance fairness in the results of graph algorithms which work 
on random walk based node embeddings.
The key idea of our method is to upweight the edges that are (1) closer to the groups' peripheries or (2) connecting different groups in the network. 
We applied our reweighting strategy
to DeepWalk and Node2Vec, 
and used the obtained representations to address fairness-enhanced influence maximization (for the original graph), node classification, and link prediction. 
Our extensive experiments confirmed the effectiveness of our algorithm to enhance fairness of various graph algorithms on
synthetic and real networks.


\appendix


\section{Classical Influence Maximization} \label{sec:appendix-classical-IM}
In the classical influence maximization problem, the goal is to find the most influential subset of $k$ nodes that can maximize the spread of a piece of information through the network. Formally, the set of $k$ most influential seeds can be found by solving the following maximization problem:
\begin{equation}\label{eq:influence}
    S^*\in{\arg\max}_{S\subseteq V} f(S),
\end{equation}
where $f:2^V\rightarrow \mathbb{R}$ quantifies the expected number of individuals infected at the end of the diffusion process.
Using IC to model information diffusion, the utility function 
$f$ in Problem \ref{eq:influence}
is a non-negative and monotone submodular set function \cite{kempe2003maximizing}. The submodularity is an intuitive notion of diminishing returns, stating that for any given sets $A \subseteq A' \subseteq V$ and any node $a \in V \setminus A'$, it holds that:
\begin{align*}
f(A \cup \{a\}) - f(A) \geq f(A' \cup \{a\}) - f(A').
\end{align*}
Although problem (\ref{eq:influence}) is NP-hard in general \cite{wolsey1982analysis}, 
for maximizing a submodular function the following greedy algorithm provides a logarithmic approximation guarantee. The greedy algorithm starts from an empty set, adds a new node to the set which provides the maximal marginal gain in terms of utility, and stops when $k$ nodes are selected.

\vspace{2mm}\noindent\textbf{Edge-reweighting Enhances Fairness}
Our reweighting method increases the weights of in-links to the nodes that are closer to the groups' peripheries. Therefore, such nodes have a higher chance of being activated in the IC model. Furthermore, our reweighting 
increases the weight of out-links from the nodes that are on the boundary of their corresponding groups. As a result, the boundary nodes activate nodes from other groups with a higher probability. This encourages the greedy seed selection algorithm to include nodes that are closer to group boundaries and can spread the information more effectively to other groups. This results in reducing the disparity of influenced individuals from various groups.

\begin{figure*}
    \begin{center}
    \includegraphics[width=.8\textwidth]{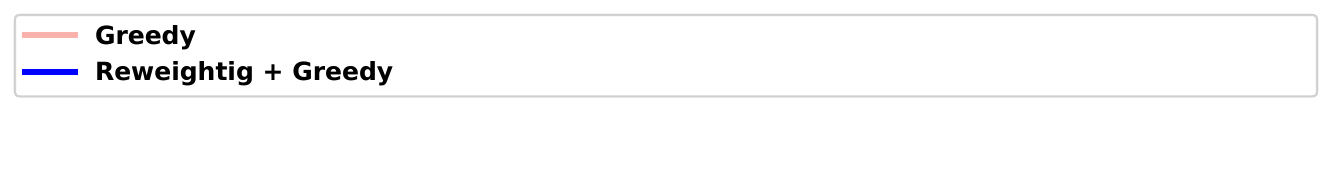}
    \\ \vspace{-1cm}
    \subfloat[\label{fig:weighted-greedy-total}]{
        \includegraphics[width=\w\textwidth]{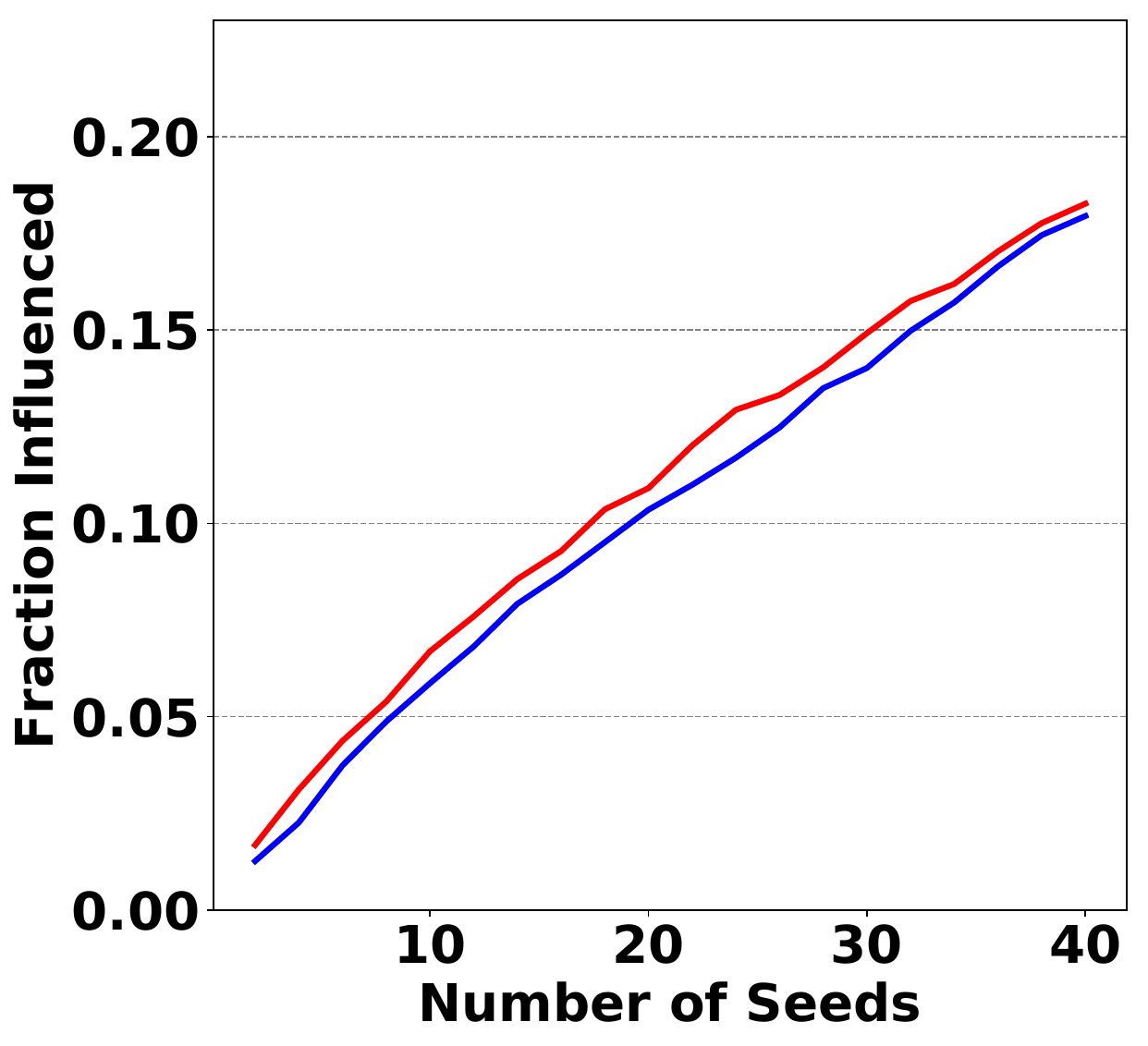}
    }
    \subfloat[\label{fig:weighted-greedy-fractions}]{
        \includegraphics[width=\w\textwidth]{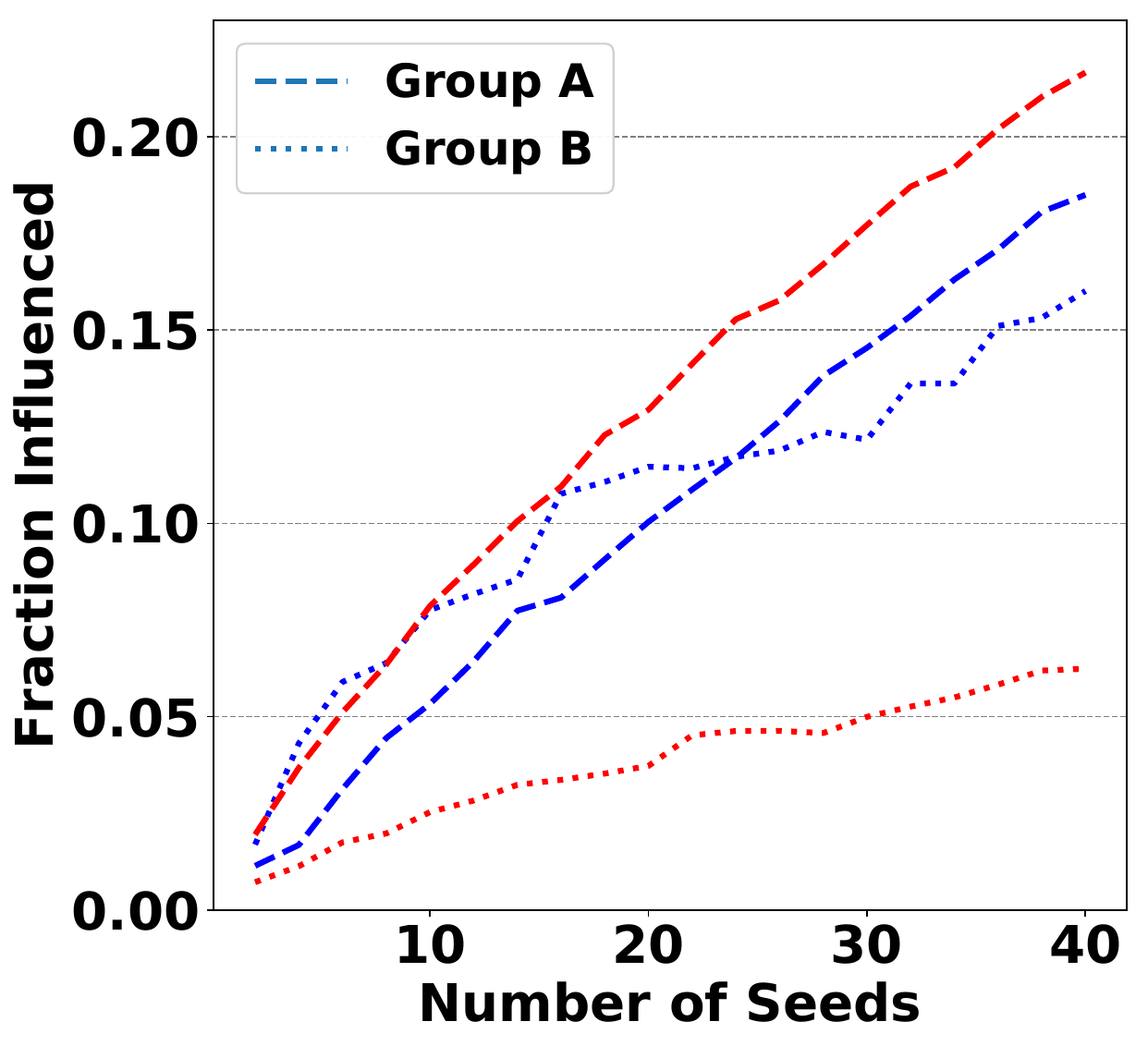}
    }
    \subfloat[\label{fig:weighted-greedy-diff}]{
        \includegraphics[width=\w\textwidth]{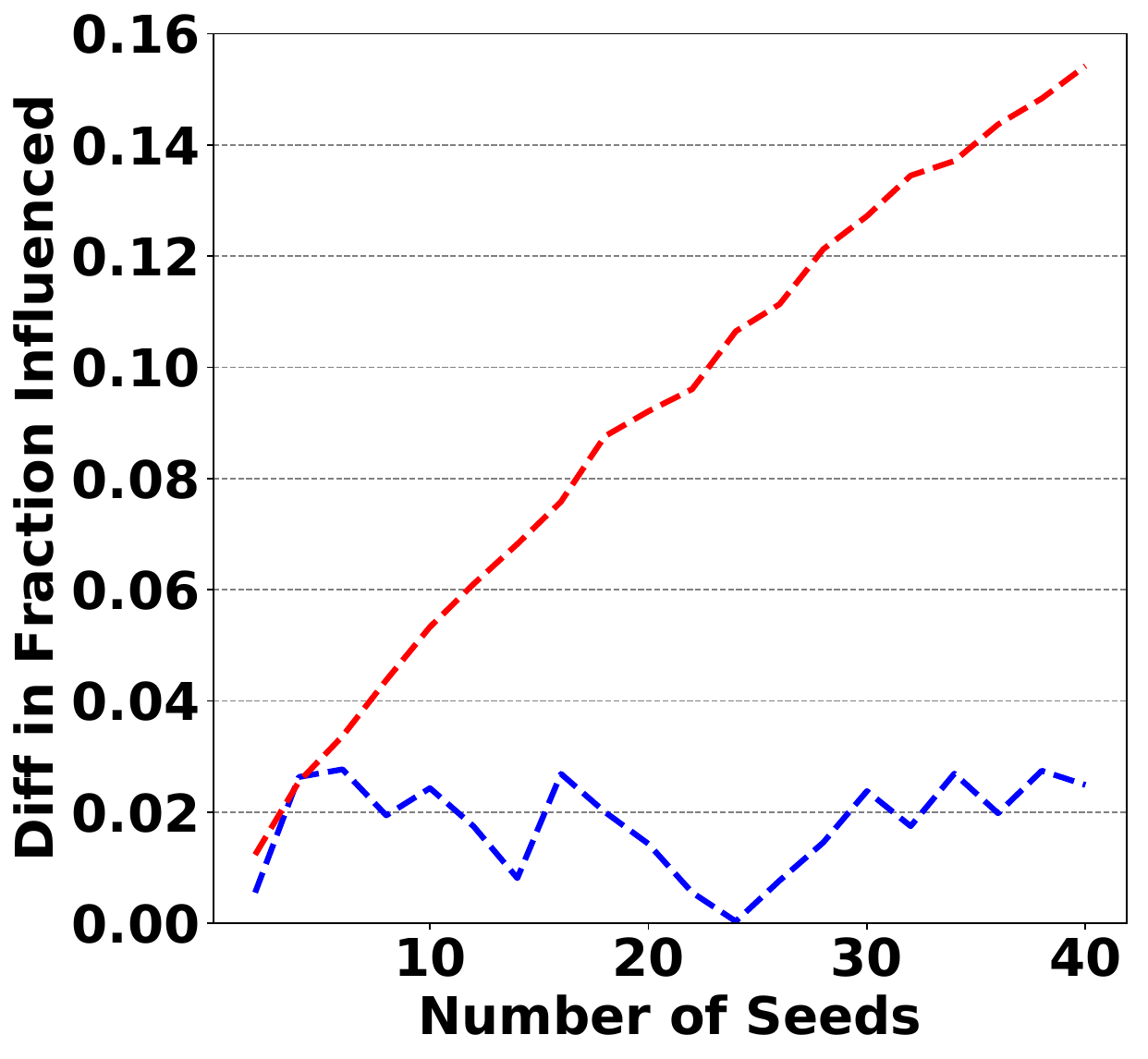}
    }
    \subfloat[\label{fig:weighted-greedy-bar}]{
        \includegraphics[width=\w\textwidth]{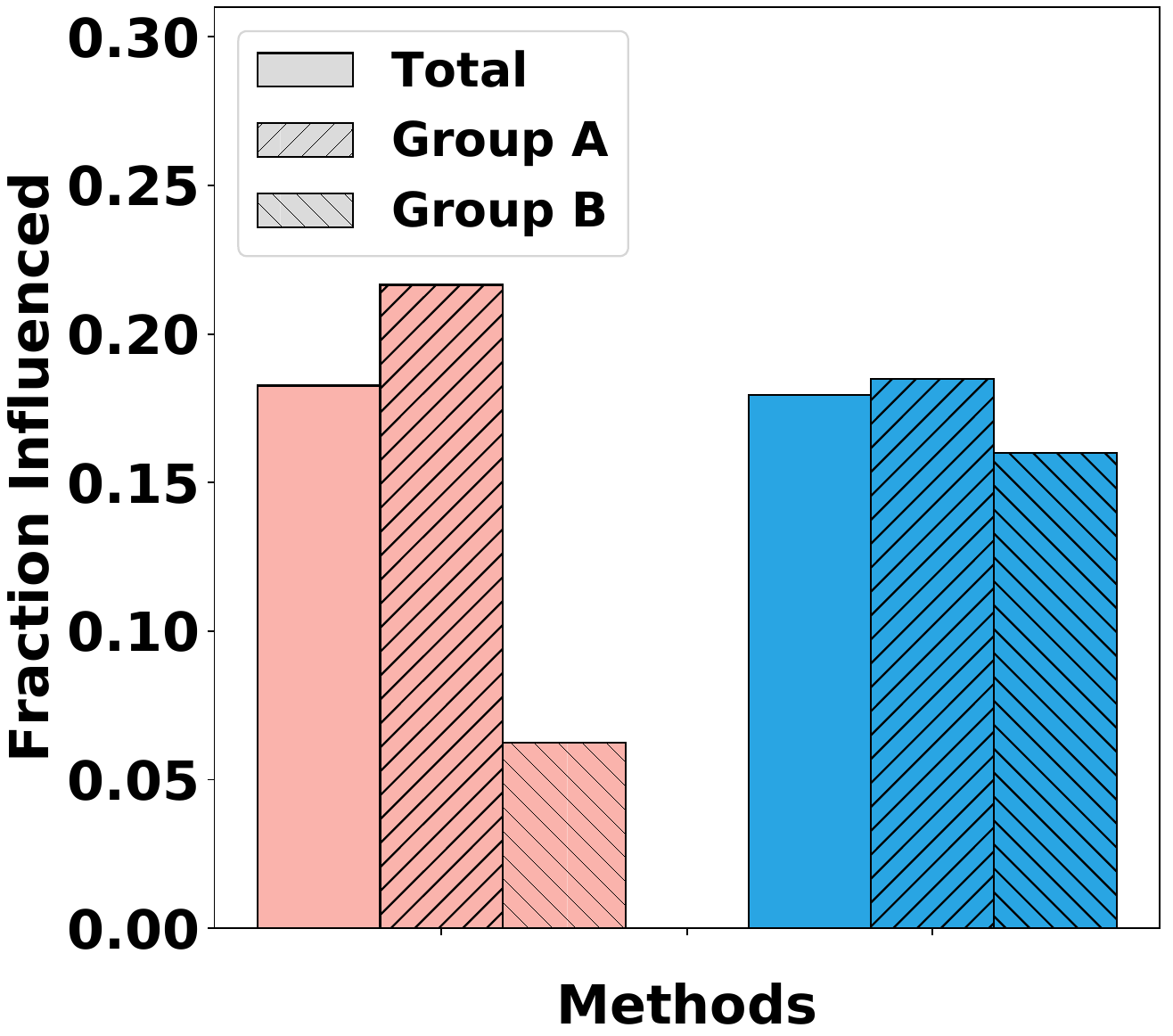}
    }
    \end{center}
    \caption{Effect of our proposed reweighting (with parameters $\alpha=0.4$ and $p=1$) on the greedy algorithm for classical influence maximization based on IC model, over the Rice-Facebook dataset. Our proposed reweighting method improves the fairness of the greedy seed selection algorithm. a) total influence. b) group-wise influence. c) influence difference between the two groups. d) total and group-wise influence.}
    \label{fig:weighted-greedy}
\end{figure*}

Figure \ref{fig:weighted-greedy} compares the fraction of influenced individuals in different groups, 
when seeds are greedily selected from a network reweighted by our proposed method. 
The greedy algorithm estimates the expected influence of each set of seeds by simulating the diffusion process based on IC model. This diffusion process can be simulated as a set of stochastic traversals of the edges on the graph. 
We scale the transmission probability of every edge $(v,u)\in E$ by the weights $w'_{vu}$ proposed by our reweighting strategy.
Formally, 
for an edge with transmission probability $\tau$ in the original graph, the scaled transmission probability can be calculated as
$\tau_{vu}' = \tau_{vu} \times w'_{vu} / \mathcal{M}_{v}$,  where $\mathcal{M}_{v} = \max_{u \in \mathcal{N}(v)} w'_{vu}$ is 
a normalization factor. It is worth mentioning that although the greedy algorithm is applied to the reweighted graph, the final evaluation of each seed selection algorithm is performed on the original graph.

\section{Effect of Bias parameters $\alpha$ and $p$} \label{sec:appendix-parameter-tuning}
To study the effect of bias parameters $\alpha$ and $p$, we use different two-grouped graphs with $P_{intra}^A = P_{intra}^B = P_{intra} = 0.025$ and different values of $P_{inter}^{AB}$. 
Figure \ref{fig:synth2Phet} shows that our proposed reweighting strategy enhances fairness of influence maximization on networks with loosely inter-connected groups (networks with smaller values of $P_{inter}^{AB}$) to a larger extent.

Comparing the results for different values of $P_{inter}^{AB}$, we see that as the number of inter-group connections grows we need to use larger values for $\alpha$. As a theoretical justification of this observation, consider a network in which a  $\beta$-fraction of neighbors of each boundary node belongs to other groups. Assume that all the edges in the original graph have equal weights.
Here, using $\alpha\geq\beta$ increases the probability of visiting nodes from other groups in random walks, but using $\alpha<\beta$ decreases the probability of crossing the groups.

The first row of Figure \ref{fig:synth2Phet} ($\alpha = 1$) shows that growing $p$ does not compensate for using small values of $\alpha$ for graphs with $P_{inter}^{AB} = 0.01$ and $P_{inter}^{AB}=0.015$. Therefore we suggest selecting larger values of $\alpha$ for graphs with larger number of inter-group connections.

When the value of $P_{inter}^{AB}$ is close to $P_{intra}^A$ and $P_{inter}^B$ ($P_{inter}^{AB} = 0.015$), the network tends to form a single connected component, 
in which each node is connected to all the other nodes with the same probability. In this settings, 
the nodes in both groups are in the same conditions, and consequently, both the greedy selection and simple DeepWalk result in fair influence maximization.

Figure \ref{fig:synth2Phet} confirms that using sufficiently large values for $p$ (larger than $2$ in this experiment) can considerably improve the resulting fairness. Note that $p$ controls the degree of bias of the stochastic edge traversal procedure towards visiting boundary nodes. 

\newcommand{\ww}{.2}
\begin{figure*}
    \begin{center}
    \includegraphics[width=.6\textwidth]{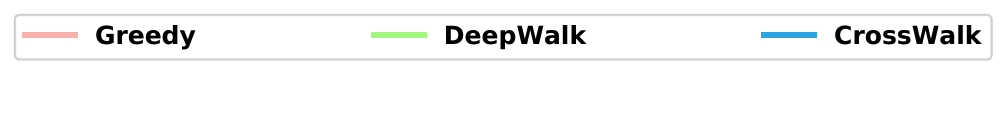}
    \begin{tabular}{cccc}
      & $p=2$ & $p=5$ & $p=8$
      \\
      \rotatebox{90}{\hspace{2cm}$\alpha=0.1$} &
      \includegraphics[width=\ww\textwidth]{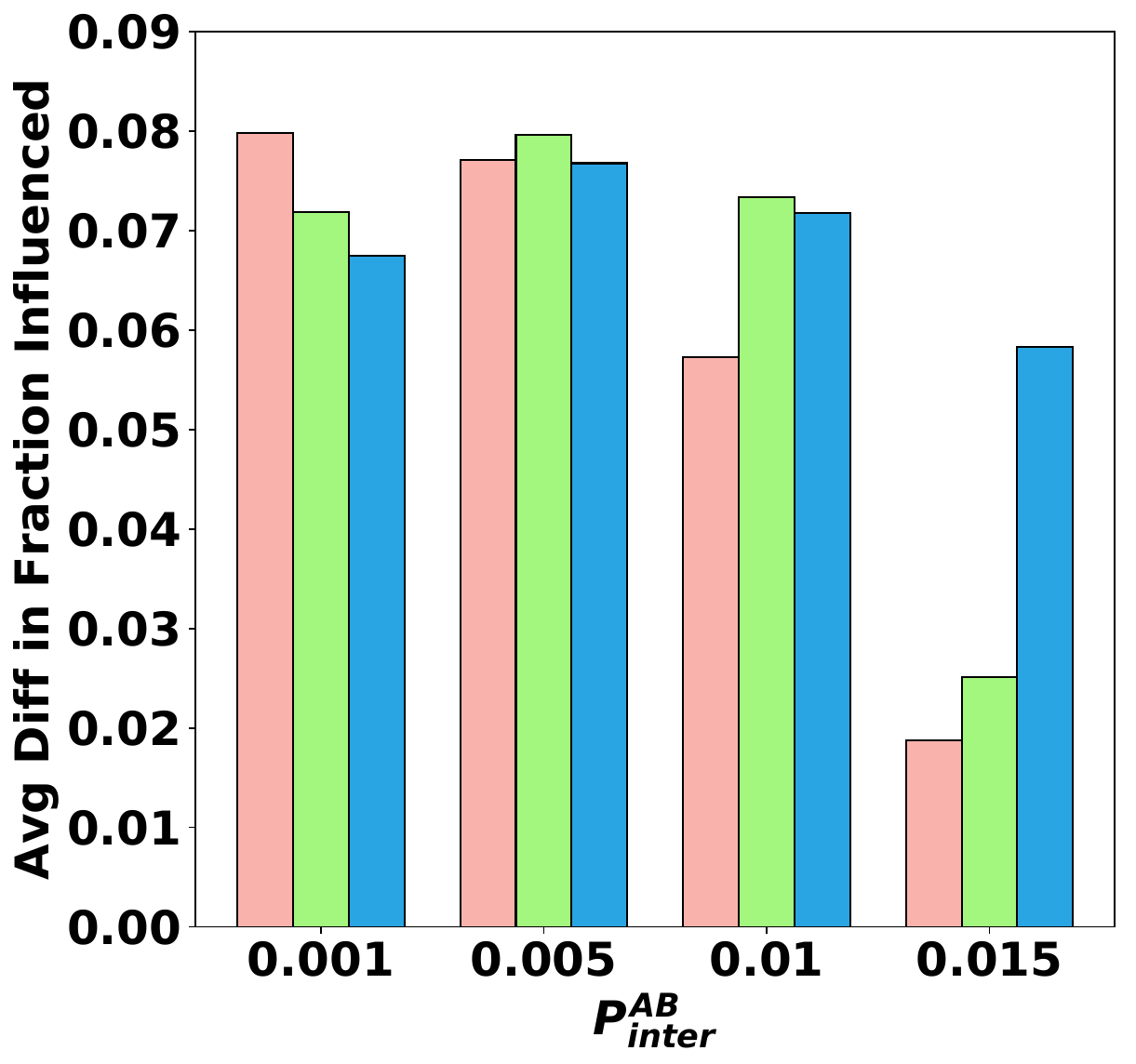}  &
      \includegraphics[width=\ww\textwidth]{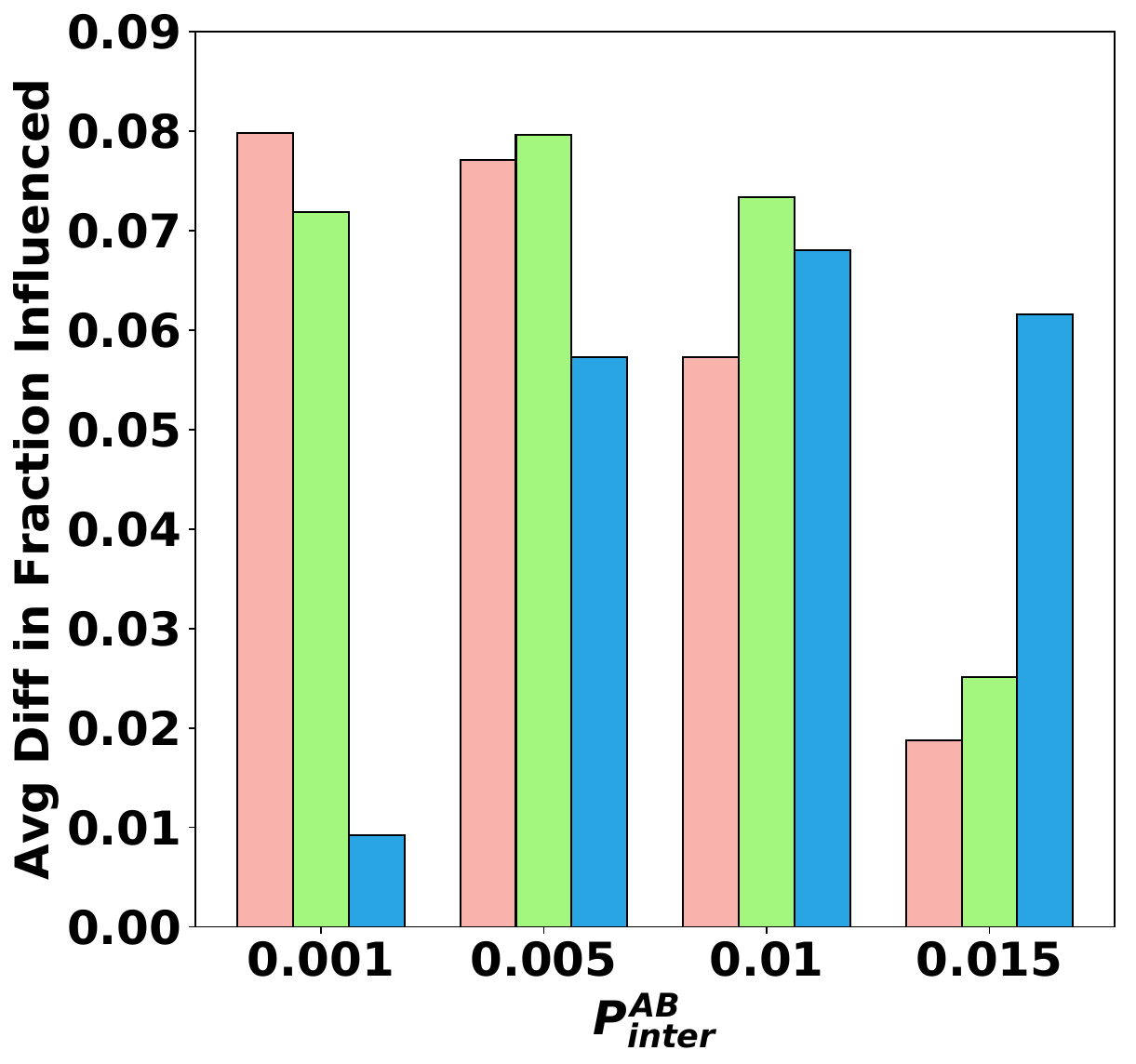} &
      \includegraphics[width=\ww\textwidth]{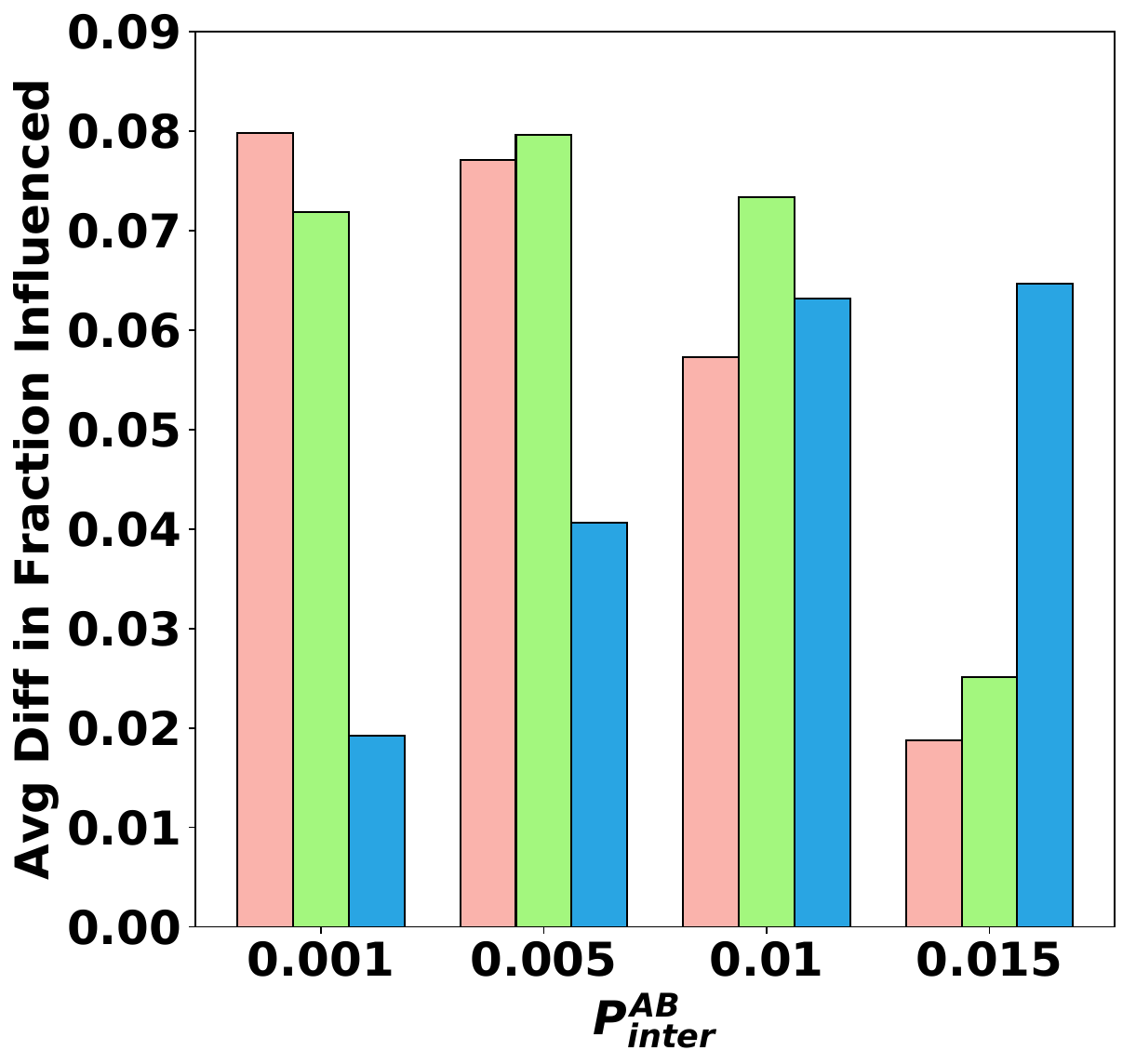}
      \\
      \rotatebox{90}{\hspace{2cm}$\alpha=0.5$} &
      \includegraphics[width=\ww\textwidth]{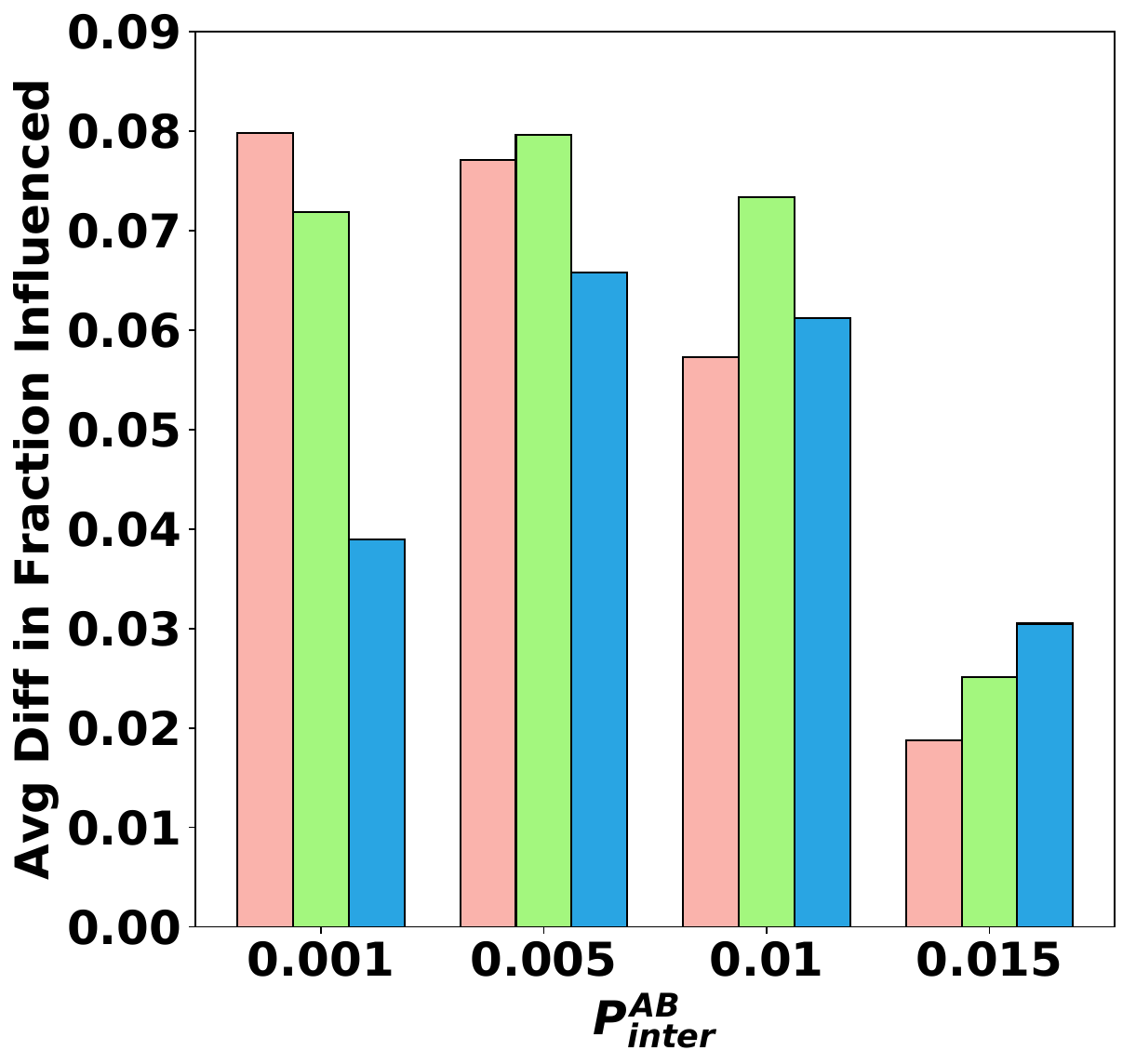}  &
      \includegraphics[width=\ww\textwidth]{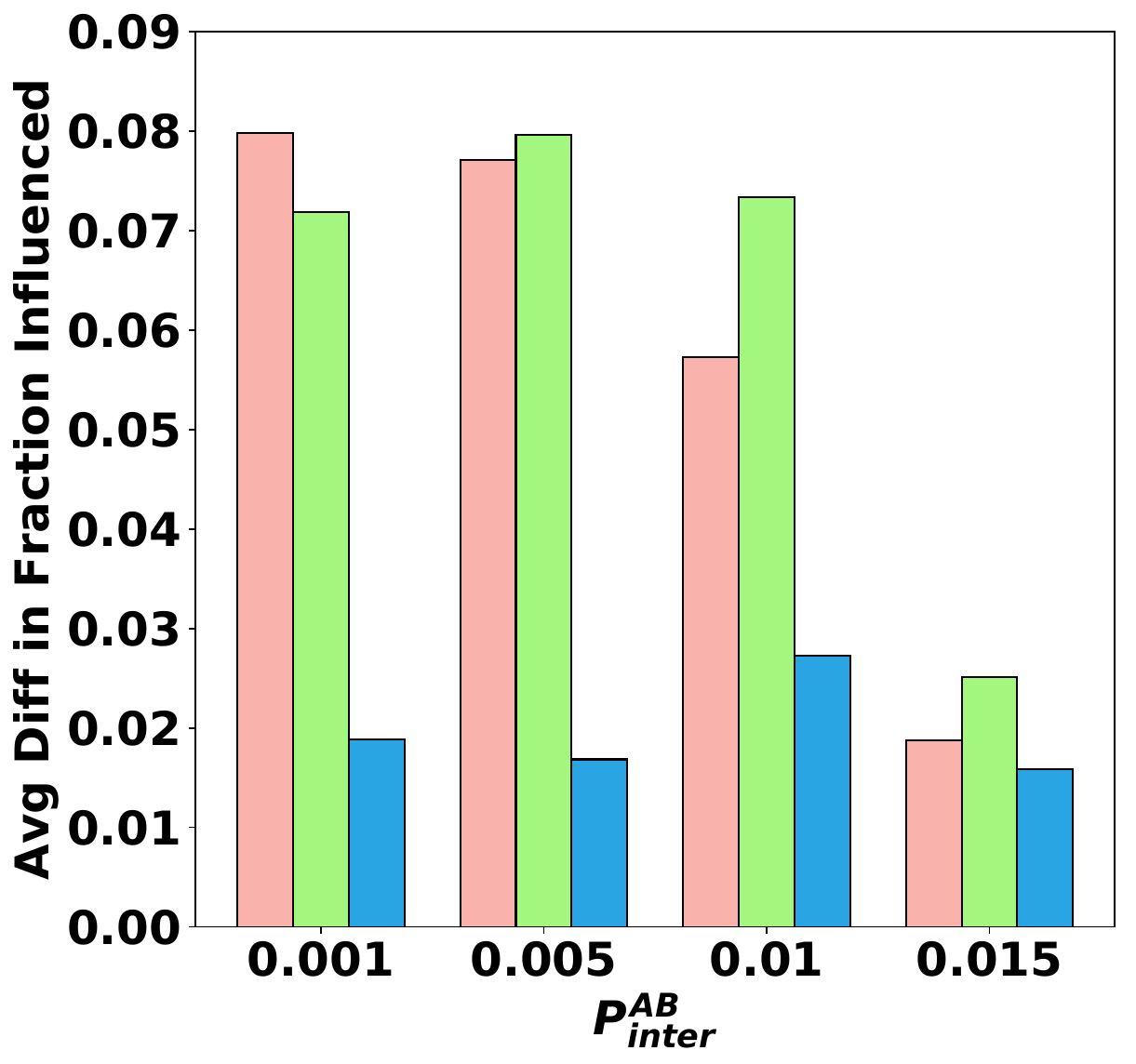} &
      \includegraphics[width=\ww\textwidth]{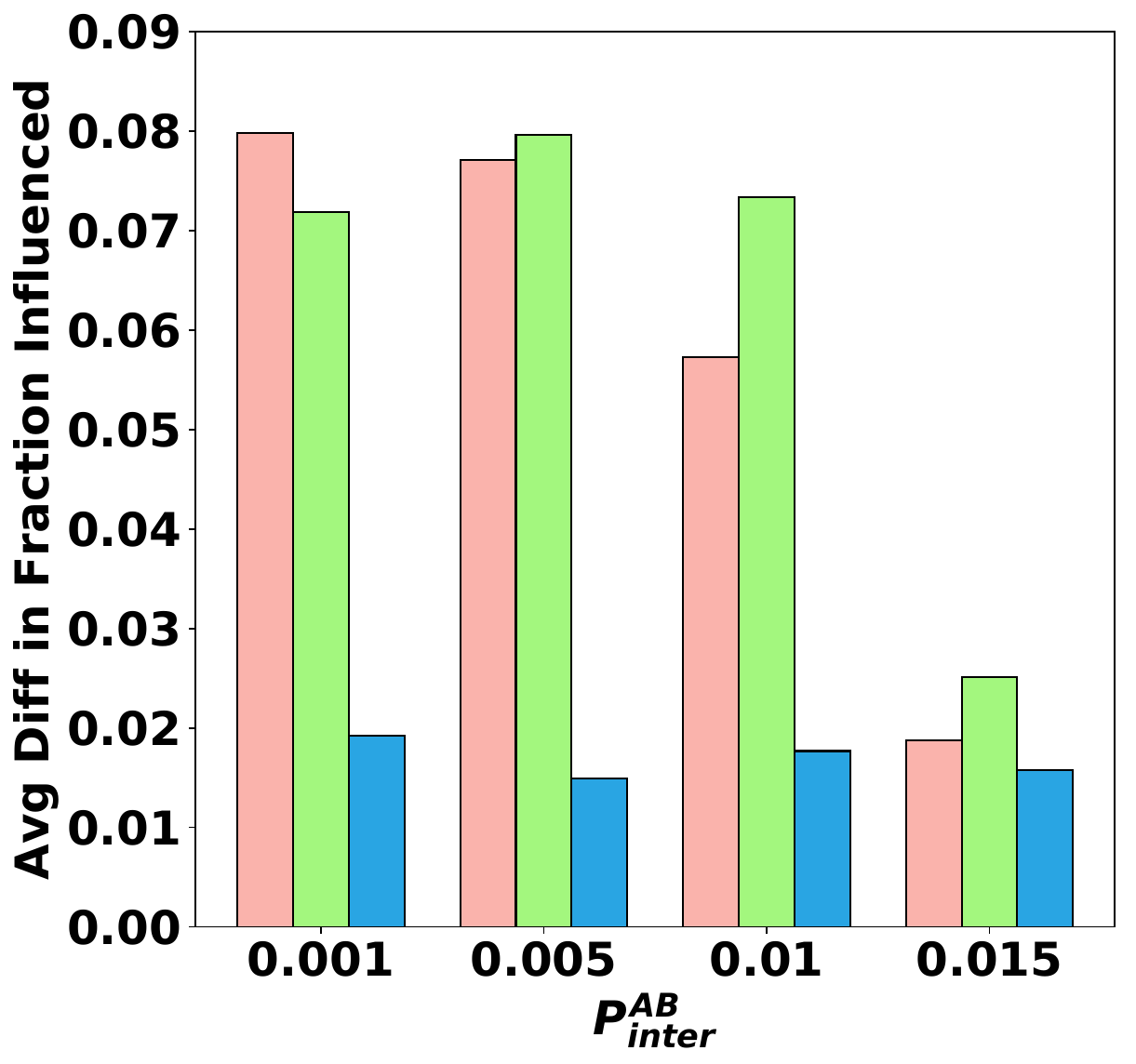}
      \\
      \rotatebox{90}{\hspace{2cm}$\alpha=0.9$} &
      \includegraphics[width=\ww\textwidth]{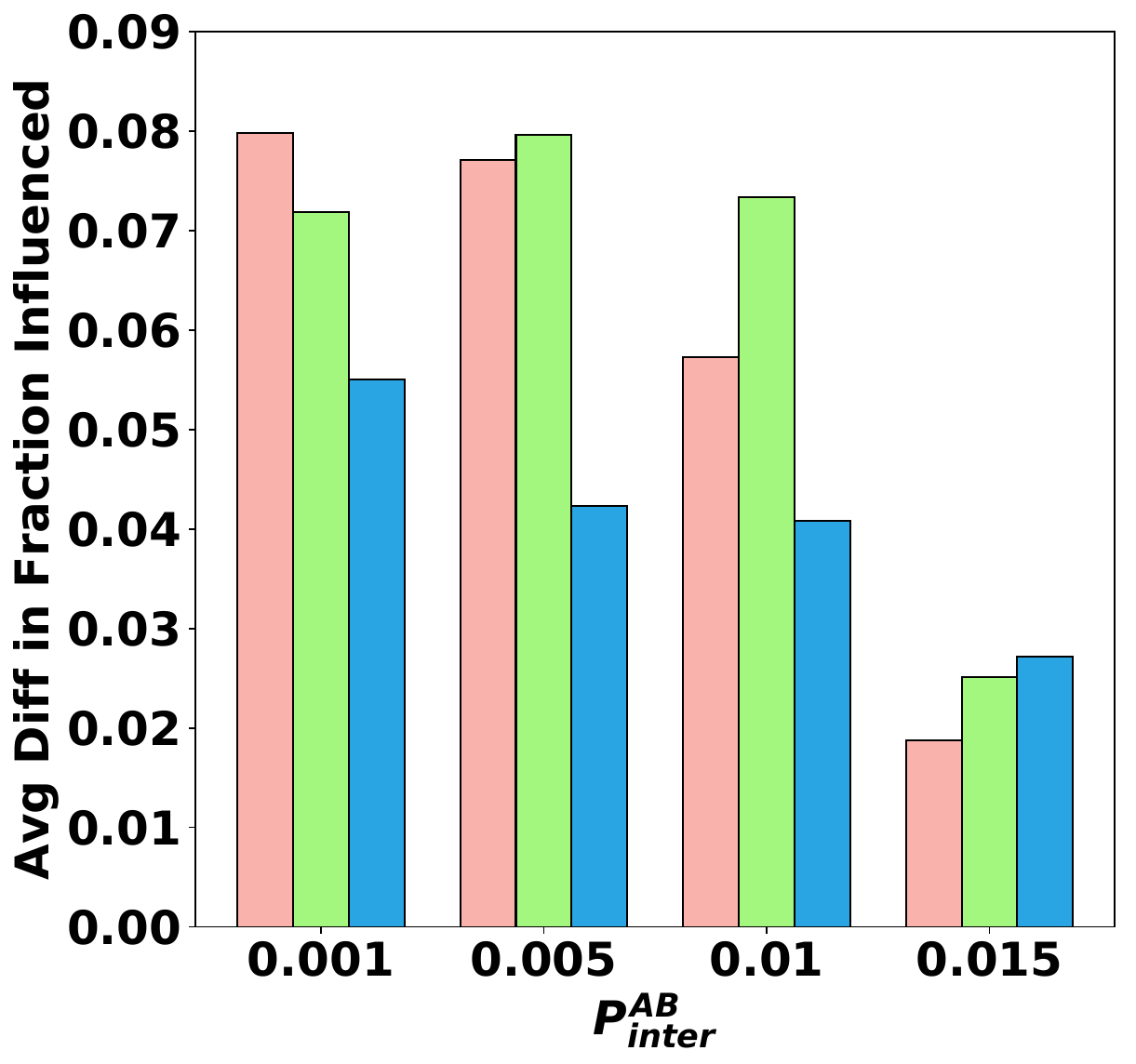}  &
      \includegraphics[width=\ww\textwidth]{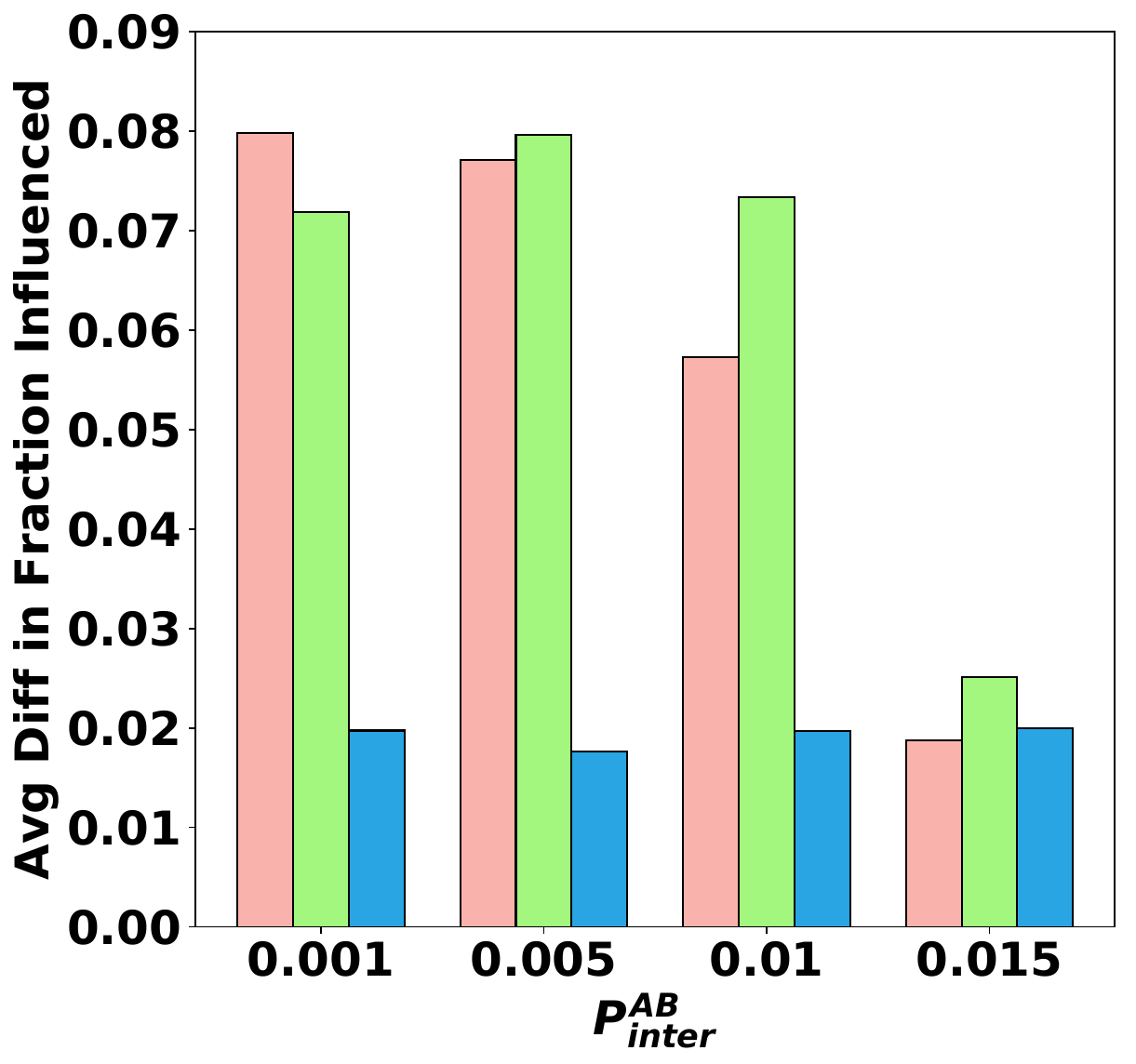} &
      \includegraphics[width=\ww\textwidth]{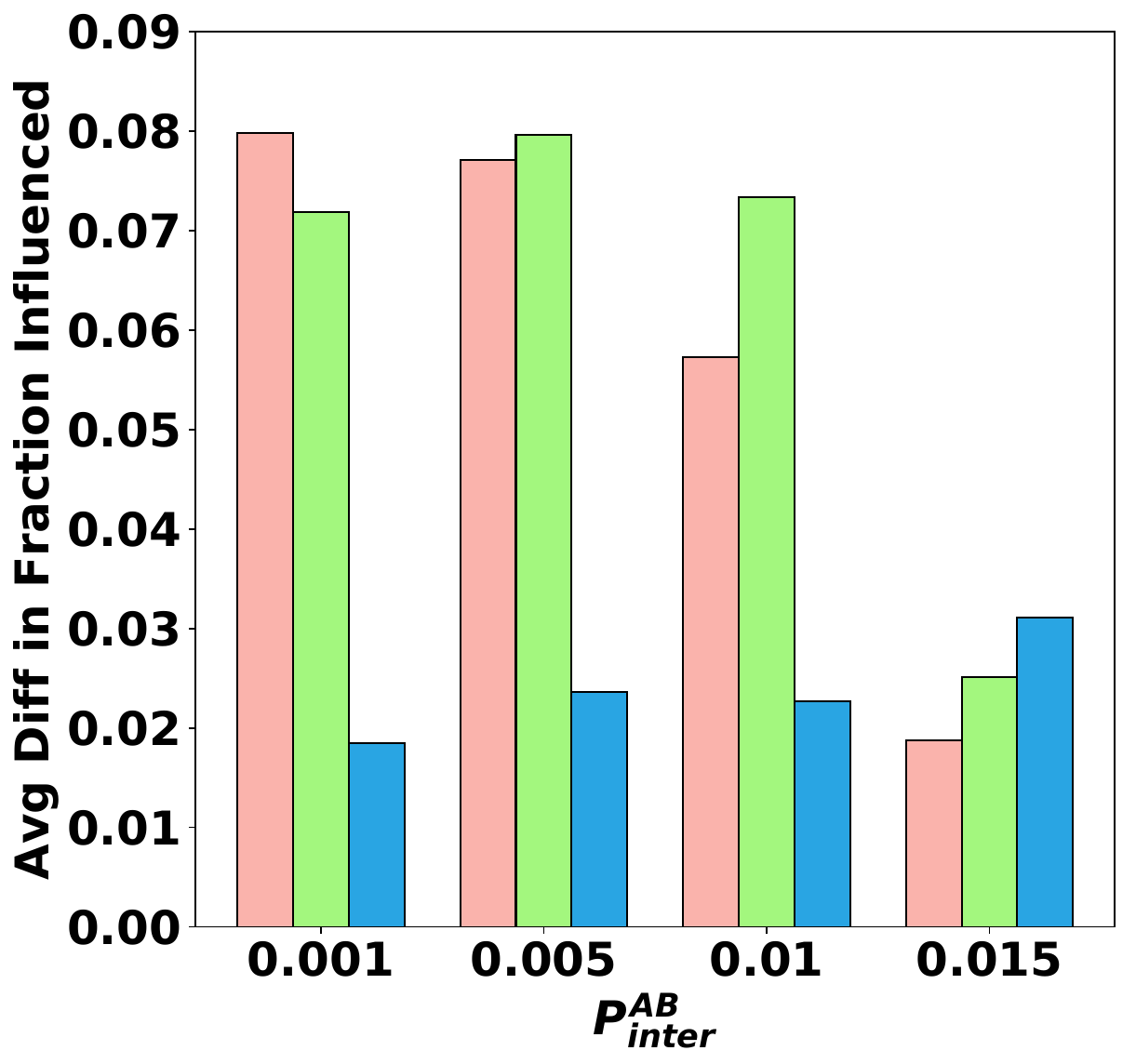}
    \end{tabular}

    \end{center}
    \caption{Effect of the parameters $\alpha$ and $p$ on the fairness of the influence, when seeding $k$-medoids of the embedings obtained by DeepWalk (green) and CrossWalk applied to DeepWalk (blue) on synthetic graphs with $P_{intra}^A = P_{intra}^B = 0.025$ and different values of $P_{inter}^{AB}$. The results of the greedy algorithm (red) are also illustrated to evaluate the fairness promotion in our method. The $y$ axis is the averaged difference in the  fraction of influenced individuals in the two groups for different values of $k$ (number of seeds) from 1 to 40.}
    \label{fig:synth2Phet}
\end{figure*}

\section{Code and Data}
Source code and datasets used in our experiments are available anonymously from \url{https://github.com/ahmadkhajehnejad/CrossWalk}.  
Datasets are located in the "data" directory.

Figures of Section \ref{sec:experiments} can be regenerated by running the jupyter notebooks in the "notebooks" directory. Figure \ref{fig:tsne} has been generated by the code in the "tsne" directory.

The "deepwalk" directory can be used to generate embeddings. Proper dataset files should be copied into the correct sub-directories before running the code.

Directories "influene\_maximization", "classifier" and "link\_prediction" contain codes for corresponding tasks. Proper dataset files and embeddings (generated by deepwalk) should be copied into the correct sub-directories before running the codes.

\newpage
\section*{Acknowledgements} 
Adrian Weller acknowledges support from a Turing AI Fellowship under grant EP/V025379/1, The Alan Turing Institute under EPSRC grant EP/N510129/1 and TU/B/000074, and the Leverhulme Trust via the CFI.
\bibliography{ref}

\begin{thebibliography}{29}
\providecommand{\natexlab}[1]{#1}

\bibitem[{Aghaei, Azizi, and Vayanos(2019)}]{aghaei2019learning}
Aghaei, S.; Azizi, M.~J.; and Vayanos, P. 2019.
\newblock Learning optimal and fair decision trees for non-discriminative
  decision-making.
\newblock In \emph{Proceedings of the AAAI Conference on Artificial
  Intelligence}, volume~33, 1418--1426.

\bibitem[{Babaei et~al.(2016)Babaei, Grabowicz, Valera, Gummadi, and
  Gomez-Rodriguez}]{babaei2016efficiency}
Babaei, M.; Grabowicz, P.; Valera, I.; Gummadi, K.~P.; and Gomez-Rodriguez, M.
  2016.
\newblock On the efficiency of the information networks in social media.
\newblock In \emph{Proceedings of the Ninth ACM International Conference on Web
  Search and Data Mining}, 83--92.

\bibitem[{Benabbou et~al.(2018)Benabbou, Chakraborty, Ho, Sliwinski, and
  Zick}]{benabbou2018diversity}
Benabbou, N.; Chakraborty, M.; Ho, X.-V.; Sliwinski, J.; and Zick, Y. 2018.
\newblock Diversity constraints in public housing allocation.
\newblock In \emph{AAMAS}, 973--981.

\bibitem[{Bose and Hamilton(2019)}]{bose2019compositional}
Bose, A.~J.; and Hamilton, W.~L. 2019.
\newblock Compositional fairness constraints for graph embeddings.
\newblock \emph{arXiv preprint arXiv:1905.10674}.

\bibitem[{Buyl and De~Bie(2020)}]{buyl2020debayes}
Buyl, M.; and De~Bie, T. 2020.
\newblock DeBayes: a Bayesian method for debiasing network embeddings.
\newblock \emph{arXiv preprint arXiv:2002.11442}.

\bibitem[{Carnes et~al.(2007)Carnes, Nagarajan, Wild, and
  Van~Zuylen}]{carnes2007maximizing}
Carnes, T.; Nagarajan, C.; Wild, S.~M.; and Van~Zuylen, A. 2007.
\newblock Maximizing influence in a competitive social network: a follower's
  perspective.
\newblock In \emph{EC}, 351--360. ACM.

\bibitem[{Cha et~al.(2010)Cha, Haddadi, Benevenuto, and Gummadi}]{cha_icwsm10}
Cha, M.; Haddadi, H.; Benevenuto, F.; and Gummadi, K.~P. 2010.
\newblock {Measuring user influence in Twitter: the million follower fallacy}.
\newblock In \emph{{Proceedings of AAAI Conference on Weblogs and Social Media
  (ICWSM'2010)}}.

\bibitem[{Easley, Kleinberg et~al.(2010)}]{easley2010networks}
Easley, D.; Kleinberg, J.; et~al. 2010.
\newblock \emph{Networks, crowds, and markets}, volume~8.
\newblock Cambridge university press Cambridge.

\bibitem[{Goyal et~al.(2013)Goyal, Bonchi, Lakshmanan, and
  Venkatasubramanian}]{goyal2013minimizing}
Goyal, A.; Bonchi, F.; Lakshmanan, L.~V.; and Venkatasubramanian, S. 2013.
\newblock On minimizing budget and time in influence propagation over social
  networks.
\newblock \emph{Social network analysis and mining}, 3(2): 179--192.

\bibitem[{Grover and Leskovec(2016)}]{grover2016node2vec}
Grover, A.; and Leskovec, J. 2016.
\newblock node2vec: Scalable feature learning for networks.
\newblock In \emph{Proceedings of the 22nd ACM SIGKDD international conference
  on Knowledge discovery and data mining}, 855--864. ACM.

\bibitem[{Hamilton, Ying, and Leskovec(2017)}]{hamilton2017inductive}
Hamilton, W.; Ying, Z.; and Leskovec, J. 2017.
\newblock Inductive representation learning on large graphs.
\newblock In \emph{Advances in Neural Information Processing Systems},
  1024--1034.

\bibitem[{Hardt, Price, and Srebro(2016)}]{hardt2016equality}
Hardt, M.; Price, E.; and Srebro, N. 2016.
\newblock Equality of opportunity in supervised learning.
\newblock \emph{Advances in neural information processing systems}, 29:
  3315--3323.

\bibitem[{Keikha et~al.(2020)Keikha, Rahgozar, Asadpour, and
  Abdollahi}]{keikha2020influence}
Keikha, M.~M.; Rahgozar, M.; Asadpour, M.; and Abdollahi, M.~F. 2020.
\newblock Influence maximization across heterogeneous interconnected networks
  based on deep learning.
\newblock \emph{Expert Systems with Applications}, 140: 112905.

\bibitem[{Kempe, Kleinberg, and Tardos(2003)}]{kempe2003maximizing}
Kempe, D.; Kleinberg, J.; and Tardos, {\'E}. 2003.
\newblock Maximizing the spread of influence through a social network.
\newblock In \emph{KDD}.

\bibitem[{Khajehnejad(2019)}]{khajehnejad2019simnet}
Khajehnejad, M. 2019.
\newblock SimNet: Similarity-based network embeddings with mean commute time.
\newblock \emph{PloS one}, 14(8): e0221172.

\bibitem[{Khajehnejad et~al.(2020)Khajehnejad, Rezaei, Babaei, Hoffmann,
  Jalili, and Weller}]{khajehnejad2020adversarial}
Khajehnejad, M.; Rezaei, A.~A.; Babaei, M.; Hoffmann, J.; Jalili, M.; and
  Weller, A. 2020.
\newblock Adversarial Graph Embeddings for Fair Influence Maximization over
  Social Networks.
\newblock \emph{arXiv preprint arXiv:2005.04074}.

\bibitem[{Kirchner and Mattu(2016)}]{kirchner2016machine}
Kirchner, J. L. L. J.~A.; and Mattu, S. 2016.
\newblock Machine Bias: There{\^a}{\u{A}}{\'Z}s Software Used Across the
  Country to Predict Future Criminals. And it{\^a}{\u{A}}{\'Z}s Biased Against
  Blacks.(May 2016).

\bibitem[{Lahoti, Gummadi, and Weikum(2019)}]{lahoti2019operationalizing}
Lahoti, P.; Gummadi, K.~P.; and Weikum, G. 2019.
\newblock Operationalizing individual fairness with pairwise fair
  representations.
\newblock \emph{arXiv preprint arXiv:1907.01439}.

\bibitem[{Masrour et~al.(2020)Masrour, Wilson, Yan, Tan, and
  Esfahanian}]{masrour2020bursting}
Masrour, F.; Wilson, T.; Yan, H.; Tan, P.-N.; and Esfahanian, A. 2020.
\newblock Bursting the Filter Bubble: Fairness-Aware Network Link Prediction.
\newblock In \emph{Proceedings of the AAAI Conference on Artificial
  Intelligence}, volume~34, 841--848.

\bibitem[{Mehrabi et~al.(2019)Mehrabi, Morstatter, Saxena, Lerman, and
  Galstyan}]{mehrabi2019survey}
Mehrabi, N.; Morstatter, F.; Saxena, N.; Lerman, K.; and Galstyan, A. 2019.
\newblock A survey on bias and fairness in machine learning.
\newblock \emph{arXiv preprint arXiv:1908.09635}.

\bibitem[{Mikolov et~al.(2013)Mikolov, Chen, Corrado, and
  Dean}]{mikolov2013efficient}
Mikolov, T.; Chen, K.; Corrado, G.; and Dean, J. 2013.
\newblock Efficient estimation of word representations in vector space.
\newblock \emph{arXiv preprint arXiv:1301.3781}.

\bibitem[{Mislove et~al.(2010)Mislove, Viswanath, Gummadi, and
  Druschel}]{mislove2010you}
Mislove, A.; Viswanath, B.; Gummadi, K.~P.; and Druschel, P. 2010.
\newblock You are who you know: inferring user profiles in online social
  networks.
\newblock In \emph{Proceedings of the third ACM international conference on Web
  search and data mining}, 251--260.

\bibitem[{Osoba and Welser~IV(2017)}]{osoba2017intelligence}
Osoba, O.~A.; and Welser~IV, W. 2017.
\newblock \emph{An intelligence in our image: The risks of bias and errors in
  artificial intelligence}.
\newblock Rand Corporation.

\bibitem[{Perozzi, Al-Rfou, and Skiena(2014)}]{deepwalk}
Perozzi, B.; Al-Rfou, R.; and Skiena, S. 2014.
\newblock Deepwalk: Online learning of social representations.
\newblock In \emph{Proceedings of the 20th ACM SIGKDD international conference
  on Knowledge discovery and data mining}, 701--710.

\bibitem[{Rahman et~al.(2019)Rahman, Surma, Backes, and
  Zhang}]{rahman2019fairwalk}
Rahman, T.~A.; Surma, B.; Backes, M.; and Zhang, Y. 2019.
\newblock Fairwalk: Towards Fair Graph Embedding.
\newblock In \emph{IJCAI}, 3289--3295.

\bibitem[{Richardson and Domingos(2002)}]{richardson2002mining}
Richardson, M.; and Domingos, P. 2002.
\newblock Mining knowledge-sharing sites for viral marketing.
\newblock In \emph{KDD}.

\bibitem[{Wang, Cui, and Zhu(2016)}]{wang2016structural}
Wang, D.; Cui, P.; and Zhu, W. 2016.
\newblock Structural deep network embedding.
\newblock In \emph{Proceedings of the 22nd ACM SIGKDD international conference
  on Knowledge discovery and data mining}, 1225--1234. ACM.

\bibitem[{Wolsey(1982)}]{wolsey1982analysis}
Wolsey, L.~A. 1982.
\newblock An analysis of the greedy algorithm for the submodular set covering
  problem.
\newblock \emph{Combinatorica}, 2(4): 385--393.

\bibitem[{Zemel et~al.(2013)Zemel, Wu, Swersky, Pitassi, and
  Dwork}]{zemel2013learning}
Zemel, R.; Wu, Y.; Swersky, K.; Pitassi, T.; and Dwork, C. 2013.
\newblock Learning fair representations.
\newblock In \emph{International Conference on Machine Learning}, 325--333.

\end{thebibliography}

\end{document}